\documentclass[3p]{elsarticle}
\usepackage{lineno,hyperref}
\usepackage{subcaption}
\usepackage{amsmath}
\usepackage{amsthm}

\usepackage{forest}
\usetikzlibrary{shapes,arrows,positioning,automata}

\tikzset{
  treenode/.style = {shape=rectangle, rounded corners,
                     draw, align=center,
                     top color=white,
                     bottom color=white},
  root/.style     = {treenode, font=\Large,
                     bottom color=red!30},
  env/.style      = {treenode, font=\ttfamily\normalsize},
  dummy/.style    = {circle,draw}
}
\usepackage{caption}
\usepackage{threeparttable}
\usepackage{algorithm,algpseudocode}
\usepackage[usestackEOL]{stackengine}
\usepackage{xcolor}
\edef\tmp{\the\baselineskip}
\setstackgap{L}{\tmp}
\usepackage{amsmath}
\modulolinenumbers[1]
\journal{Journal}

\begin{document}

\begin{frontmatter}

\title{Active Learning with Multifidelity Modeling for Efficient Rare Event Simulation}

\author[A1]{Somayajulu L. N. Dhulipala\corref{mycorrespondingauthor}}
\address[A1]{Computational Mechanics and Materials, Idaho National Laboratory, Idaho Falls, ID 83402, USA}\cortext[mycorrespondingauthor]{Corresponding author; Email: Som.Dhulipala@inl.gov}
\author[A2]{Michael D. Shields}
\address[A2]{Department of Civil and Systems Engineering, Johns Hopkins University, Baltimore, MD 21218, USA}
\author[A1]{Benjamin W. Spencer}
\author[A3]{Chandrakanth Bolisetti}
\address[A3]{Advanced Reactor Technology and Design, Idaho National Laboratory, Idaho Falls, ID 83402, USA}
\author[A4]{Andrew E. Slaughter}
\address[A4]{Computational Frameworks, Idaho National Laboratory, Idaho Falls, ID 83402, USA}
\author[A5]{Vincent M. Labour\'e}
\address[A5]{Reactor Physics Methods and Analysis, Idaho National Laboratory, Idaho Falls, ID 83402, USA}
\author[A2]{Promit Chakroborty}

\begin{abstract}
While multifidelity modeling provides a cost-effective way to conduct uncertainty quantification with computationally expensive models, much greater efficiency can be achieved by adaptively deciding the number of required high-fidelity (HF) simulations, depending on the type and complexity of the problem and the desired accuracy in the results. We propose a framework for active learning with multifidelity modeling emphasizing the efficient estimation of rare events. Our framework works by fusing a low-fidelity (LF) prediction with an HF-inferred correction, filtering the corrected LF prediction to decide whether to call the high-fidelity model, and for enhanced subsequent accuracy, adapting the correction for the LF prediction after every HF model call. The framework does not make any assumptions as to the LF model type or its correlations with the HF model. In addition, for improved robustness when estimating smaller failure probabilities, we propose using dynamic active learning functions that decide when to call the HF model. We demonstrate our framework using several academic case studies and two finite element (FE) model case studies: estimating Navier-Stokes velocities using the Stokes approximation and estimating stresses in a transversely isotropic model subjected to displacements via a coarsely meshed isotropic model. Across these case studies, not only did the proposed framework estimate the failure probabilities accurately, but compared with either Monte Carlo or a standard variance reduction method, it also required only a small fraction of the calls to the HF model.
\end{abstract}


\end{frontmatter}


\section{Introduction}

Multifidelity modeling substitutes and/or {augments} ``exact'' but computationally expensive high-fidelity (HF) models with cheaper but approximate low-fidelity (LF) models \cite{Perdikaris2016z,Giselle2019z,Guo2018z}. This modeling strategy is finding many uses in computational sciences and engineering; consequently, recent research has focused on more effective and efficient approaches for multifidelity modeling in uncertainty quantification and propagation \cite{Teckentrup2015a}, optimization \cite{Li2020a}, and inverse analysis \cite{Gorodetsky2020z}. Monte Carlo {simulation}, which typically requires numerous evaluations of an HF model, can be considerably accelerated through multifidelity modeling strategies \cite{Zhang2020a}. {Of particular interest are rare events associated with small failure probabilities that are difficult to estimate and are important across multiple applications (e.g., aerospace systems reliability \cite{Morio2015a}, critical infrastructure resilience to natural hazards \cite{Zio2021a}, and advanced nuclear fuel safety \cite{Jiang2021a}).} To efficiently estimate the likelihood of rare events, we propose a framework for active learning with multifidelity modeling.

{
\subsection{Brief Review of multifidelity modeling and active learning for reliability}}

\citet{Peherstorfer2018a} broadly classifies multifidelity modeling strategies into three categories: \textit{fusion}, which combines information from HF and LF models; \textit{adaptation}, which corrects the LF model after each evaluation or set of evaluations of the HF model; and \textit{filtering}, which decides whether to call the HF model only after calling LF models first. In Monte Carlo {simulation}, multifidelity modeling using {fusion} has been a popular approach for fast, cost-effective estimation of the output statistics \cite{Qian2018a,Quaglino2019a}. \citet{Peherstorfer2016y} and \citet{Kramer2019a} presented a fusion of multiple models in an importance sampling scheme for the efficient estimation of rare events. {Their approach relies on finding an adequate number of samples in the failure region across multiple models in order to accurately characterize the biasing densities; it also requires the analyst to specify a priori the number of HF model calls. For smaller failure probabilities (i.e., on the order $1\times10^{-4}$ or less) and/or complex failure boundaries, these two requirements may constrain their method's performance.} \citet{Yang2019a,Yi2021a} apply co-kriging to combine information from multiple models by using the linear correlations between these models. \citet{Perdikaris2017a} points out that relying on linear correlations between HF and LF models may lead to erroneous estimation of the output statistics when used outside the validity range, and proposed an approach to consider nonlinear correlations between these models. Control variates is another popular approach for the fusion of information from multiple models, and \citet{Gorodetsky2020a} proposed an approximate control variates framework for handling multiple modeling fidelities with unknown statistics. \citet{Pham2021a} apply approximate control variates to the problem of rare events estimation in a multifidelity importance sampling scheme. {While their approach enhances variance reduction due to the consideration of correlations among the modeling fidelities, it may still face the same issues (e.g., accurate characterization of the biasing distribution and fixing the number of HF calls a priori.)} Other recent contributions have also used {fusion} for estimating HF model responses in a deterministic setting: \citet{Ahmed2021a} propose a zonal multifidelity modeling framework, \citet{Hebbal2021a} use a deep Gaussian process ($\mathcal{GP}$) to handle input parameter incoherences across the multiple models, and \citet{Meng2021a} propose a Bayesian neural network to link together a data-driven deep neural network (DNN) and a physics-informed neural network (PINN).

{Filtering} is another effective approach to multifidelity modeling, as it automatically decides when to call an HF model, and, in a Monte Carlo scheme, relies on LF models most of the time. Delayed-rejection-type Markov chain Monte Carlo (MCMC) schemes provide a framework for performing filtering by calling the HF model only when LF-based proposals are rejected \cite{Prescott2020y}. \citet{Catanach2020y} propose a multifidelity sequential tempered MCMC sampler and apply it to a chemical kinetics problem. Using {adaptation}, \citet{Nabianh2021y} propose a PINN for MCMC sampling that is refined on the fly. There have also been a combination of alternative multifidelity modeling strategies. \citet{Chakraborty2021a} uses fusion and adaptation by proposing a transfer-learning-based PINN. \citet{Zhang2018y} combines adaptation and filtering in a MCMC scheme by using an adaptive $\mathcal{GP}$ and calling the HF model only in regions of high posterior densities; they apply their framework to the inverse uncertainty quantification of a hydrologic system. An adaptive approach that refines the LF model, decides when to call the HF model, and learns the failure boundary on the fly can provide flexibility and robustness for rare events estimation using multiple models, while also significantly reducing computational costs.

Adaptive approaches using active learning have become popular in the reliability estimation literature, {although most rely} {on a single-model fidelity (i.e., only the HF model)}. \citet{Echard2011a} use two active learning functions based on $\mathcal{GP}$---namely, the $U$-function and the expected feasibility function \cite{bichon2008efficient}---to decide when to call the model in a Monte Carlo scheme. \citet{Lelievre2018a,Haj2021a} propose improved active learning functions for efficient Monte Carlo estimation aimed at reducing the number of calls to the model. \citet{Razaaly2020a} propose using an isotropic Gaussian importance sampling density to extend active-learning-based Monte Carlo for smaller failure probabilities. They also point out that this importance sampling density, due to its restrictive assumptions, can face issues in reagrd to high-dimensional spaces and nonlinear limit state functions. Active learning has also been used in Monte Carlo schemes with variance reduction for handling smaller failure probabilities in an effort to make few to no assumptions about the complexity of the failure domain. For example, \citet{Huang2016a,Zhang2019a,Xu2020a} use active learning in a subset simulation algorithm \cite{Au2001a}, and \citet{Yang2020b} use active learning in an importance sampling algorithm. \citet{Cui2019a}, however, point out that these methods may lose accuracy under complex failure domains and smaller failure probabilities. Generally speaking, in active learning, static active learning functions that decide when to call the model can break down under smaller failure probabilities, due to large differences between the nominal model outputs and the required failure threshold. Moreover, most algorithms in the reliability estimation literature have a training phase in which a large number of $\mathcal{GP}$ predictions must be made. Since the computational complexity of a $\mathcal{GP}$ for prediction and uncertainty quantification is $\mathcal{O}(dMN)$ and $\mathcal{O}(MN^2)$, respectively (where $d$ is the number of parameters dimensionality; $N$ is training set size; and $M$ is test set size) \cite{Raykar2007}, such a training phase can become a bottleneck for problems with smaller failure probabilities.

{
\subsection{Problem statement and overview of the proposed solution}}

\begin{figure}[h]
\centering
\includegraphics[width=5.7in, height=3.18in]{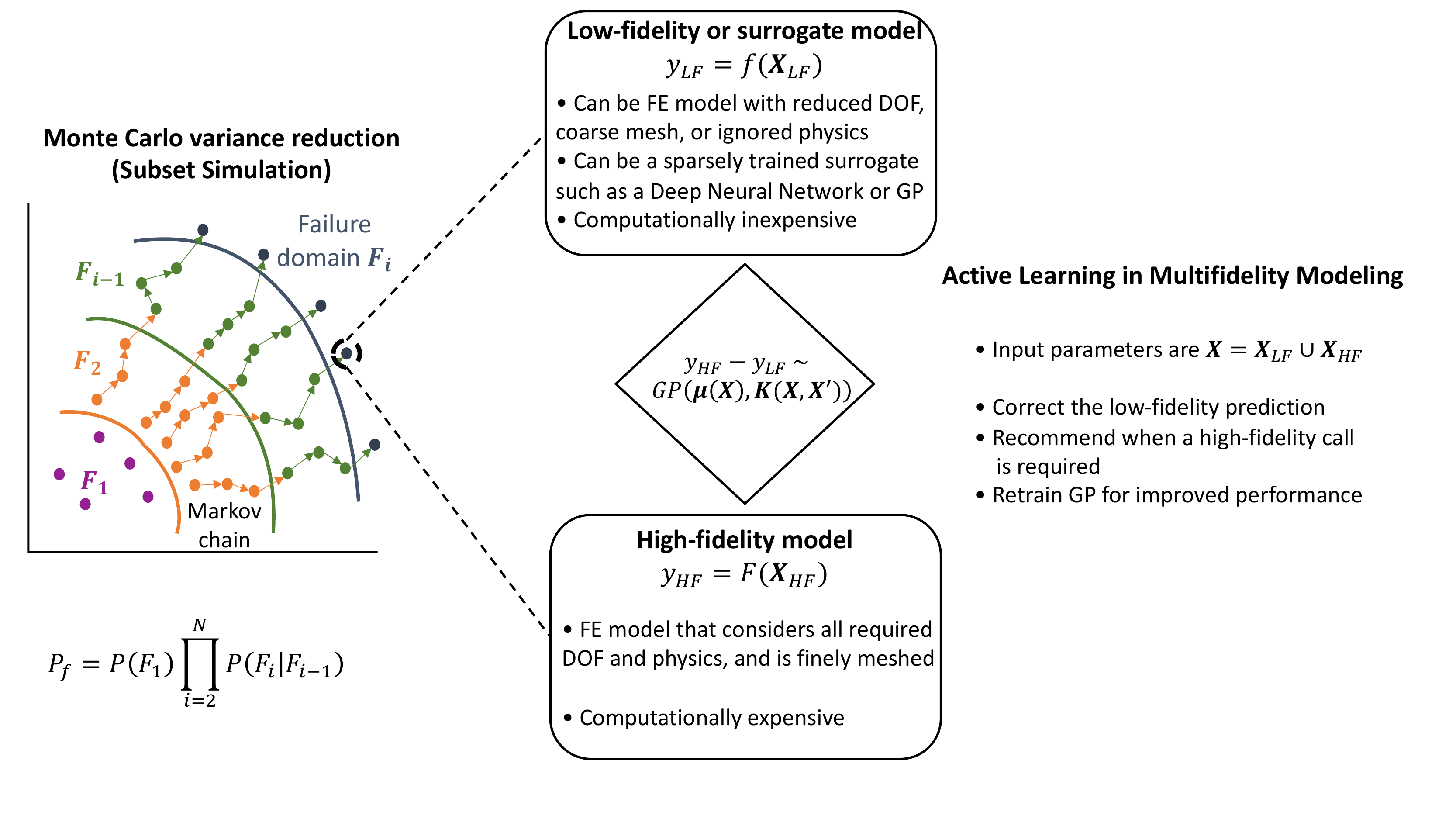} 
\caption{Schematic of the proposed framework for active learning with multifidelity modeling. This framework relies on subset simulation for variance reduction and uses fusion, filtering, and adaptation, respectively, to correct the low-fidelity predictions via a Gaussian process, decide when to call the high-fidelity model, and retrain the Gaussian process if the high-fidelity model is called. The mathematical definition of a Gaussian process is presented in Eq.\ \eqref{eqn:GP_f}-\eqref{eqn:GP_likelihood}.}
\label{Schematic}
\end{figure}

Rare events characterization involves computing the following integral {to compute the probability of failure}:

\begin{equation}
    \label{eqn:Pf_exact}
    P_f = \int_{\widetilde{F}(\pmb{X}) > \mathcal{F}} q(\pmb{X})~d\pmb{X} 
\end{equation}

\noindent where $\pmb{X}$ is the vector of random input model parameters, $q(.)$ is their probability {density function}, $\widetilde{F}(\pmb{X})$ is the required model prediction, and $\mathcal{F}$ is the failure threshold. {For most applications, t}he above integral is intractable to solve in closed form, owing to its dimensionality and the complexity of {the} failure boundary defined by $\widetilde{F}(\pmb{X}) > \mathcal{F}$. A Monte Carlo estimator {for} the above integral is:

\begin{equation}
    \label{eqn:Pf_approx}
    P_f \approx \hat{P}_f = \frac{1}{N_m}~\sum \mathbf{I}\big(\widetilde{F}(\pmb{X}) > \mathcal{F}\big)
\end{equation}

\noindent where $N_m$ is the number of Monte Carlo samples and $I(.)$ is an indicator function. In standard Monte Carlo, a large number of HF model evaluations must be made to estimate $P_f$ accurately. Our framework uses {three steps: 1. a fusion step; 2. a filtering step; and 3. an adaptation step, within a subset simulation Monte Carlo method to achieve variance reduction and  leverage multi-fidelity models for $P_f$ estimation}. As presented in Figure \ref{Schematic} {and described in Section \ref{sec:SS}}, subset simulation operates by creating intermediate failure thresholds (i.e., expressing small $P_f$ values as a product of larger intermediate failure probabilities) and simulating a number of Markov chains that {propagate} to the failure region, while making no assumptions as to its complexity. For each model evaluation in subset simulation, our framework first evaluates an LF model and then adds a $\mathcal{GP}$ correction term inferred from previous HF model calls. This is the fusion step. Next, the decision is made on whether or not to call the HF model. This is the filtering step, and is based on dynamic active learning functions. Finally, if an HF call is made, the $\mathcal{GP}$ (which provides a correction to the LF predictions) is updated with this new information. This is the adaptation step.

{\subsection{Contributions of this work}}
{The proposed} framework fuses the LF prediction with a HF-inferred $\mathcal{GP}$ correction, filters the LF prediction to decide whether to call the HF model, and, for enhanced accuracy of subsequent corrections, adapts the LF correction for every HF call. In doing so, it makes the following primary contributions:

\begin{itemize}
    {\item The proposed framework leverages the LF model(s) to estimate $P_f$ with only a small number of HF model evaluations.}
    \item {The proposed method provides flexibility} in the choice of LF model by not making any assumptions as to model type (i.e., surrogate, reduced physics, or reduced degrees of freedom [DoFs]) or correlations with the HF model. The LF model is also allowed to operate on a different set of input parameters than the HF model.
    \item {The proposed method employs} dynamic active learning functions that evolve as the algorithm proceeds, thus deciding when to call the HF model in a way that ensures that active learning does not break down for smaller failure probabilities.
    \item  A $\mathcal{GP}$ is applied, with the test size (or the number of samples to be evaluated) always being one as the algorithm proceeds. Therefore, the computational complexity is $\mathcal{O}(dN)$ and $\mathcal{O}(N^2)$ for prediction and uncertainty quantification, respectively. Additionally, the $\mathcal{GP}$ is trained only when a HF model is called; therefore, the training set size will be a very small fraction of the total number of samples evaluated. 
\end{itemize}

\noindent 
We demonstrate the proposed framework on several academic case studies and two finite element (FE) model case studies. Also, the notations used in this paper are defined in Appendix A.

\section{Background}

In this section, we briefly review $\mathcal{GP}$ {regression (Kriging)} and subset simulation, {and propose an active learning approach for HF versus LF model selection within subset simulation.}

\subsection{Gaussian Process Regression}

A function $f(\pmb{X})$ is said to be a $\mathcal{GP}$ if it follows a {joint} normal distribution with mean and covariance functions ${m}(\pmb{X})$ and ${k}(\pmb{X},~\pmb{X}^\prime)$, respectively \cite{Rasmussen2004}:



\begin{equation}
    \label{eqn:GP_f}
    f(\pmb{X}) \sim \mathcal{N}\big(m(\pmb{X}), k(\pmb{X},\pmb{X}^\prime)\big)
\end{equation}




\noindent Given {the general flexibility for a $\mathcal{GP}$ to model the relation between input-output data, $\mathcal{GP}$s are often used as surrogate models to predict new output values at previously unsampled input values. That is, given some training data $\{\pmb{X}, \pmb{y}\}$, a $\mathcal{GP}$ can be used to make a prediction at of the output $\pmb{y}_*$ at a new input value $\pmb{X}_*$, by exploiting the joint Gaussian distribution between the training data and the new sample points, i.e.}



\begin{equation}
    \label{eqn:GP_prior}
    \begin{Bmatrix}
    \pmb{y}\\
    \pmb{y}_*
    \end{Bmatrix} \sim \mathcal{N}\Bigg(\pmb{0}, \begin{bmatrix}
    k(\pmb{X},\pmb{X}) & k(\pmb{X},\pmb{X}_*)\\
    k(\pmb{X}_*,\pmb{X}) & k(\pmb{X}_*,\pmb{X}_*)
    \end{bmatrix}\Bigg)	
\end{equation}

\noindent 
{In a Bayesian framework, the} posterior predictive distribution of $\pmb{y}_*$, given the training/new inputs and the training outputs, is:

\begin{equation}
    \label{eqn:GP_posterior}
    \begin{aligned}
    p(\pmb{y}_*~|~\pmb{X}, \pmb{X}_*, \pmb{y}) \sim \mathcal{N}\Big(~ & k(\pmb{X}_*,\pmb{X})~k(\pmb{X},\pmb{X})^{-1}~\pmb{y}, \\
    & k(\pmb{X}_*,\pmb{X}_*) - k(\pmb{X}_*,\pmb{X})~k(\pmb{X},\pmb{X})^{-1}~k(\pmb{X},\pmb{X}_*)~\Big)
    \end{aligned}
\end{equation}

\noindent {To determine the precise mean and variance of $\pmb{y}_*$, it is necessary to infer/learn a set of hyperparameters for the covariance function ${k}(\pmb{X},~\pmb{X}^\prime)$.} This {parameter learning is often} accomplished by minimizing the negative marginal log-likelihood $\mathcal{L}$ with respect to the hyperparameters:

\begin{equation}
    \label{eqn:GP_likelihood}
    \mathcal{L} = -\ln~p(\pmb{y}~|~\pmb{X},\sigma^2,\lambda) \propto \frac{1}{2}~\ln |k(\pmb{X},\pmb{X})| + \frac{1}{2}~\pmb{y}^T~k(\pmb{X},\pmb{X})^{-1}~\pmb{y}
\end{equation}

\noindent where the marginal likelihood $p(\pmb{y}~|~\pmb{X},\sigma^2,\lambda)$ follows a normal distribution.

\subsection{Monte Carlo variance reduction with subset simulation}\label{sec:SS}

Subset simulation is a variance reduction framework proposed by \citet{Au2001a} for estimating small failure probabilities in high-dimensional spaces. This framework operates on the principal of expressing a small failure probability as a product of {larger (and thereby easier to estimate)} intermediate failure probabilities:

\begin{equation}
    \label{eqn:SS1}
    P_f = P(\widetilde{F}(\pmb{X})>\mathcal{F}_1) \prod_{s=2}^{N_s}~P(\widetilde{F}(\pmb{X})>\mathcal{F}_s|\widetilde{F}(\pmb{X})>\mathcal{F}_{s-1}) \equiv P_1 \prod_{s=2}^N~P_{s|s-1} %
\end{equation}

\noindent where $P_1$ and $P_{s|s-1}$ are intermediate failure probabilities of the first and subsequent subsets, respectively. 
While Monte Carlo is used to estimate the probability $P_1$, {it is generally necessary to use Markov Chain Monte Carlo (MCMC) methods to sample from the conditional densities in each subset and estimate the conditional probabilities} $P_{s|s-1}$. \citet{Au2001a} originally proposed a component-wise Metropolis-Hastings algorithm to estimate $P_{s|s-1}$. Recently, other MCMC methods such as delayed rejection \cite{Papaioannou2015a}, Hamiltonian Monte Carlo \cite{Wang2019a}, and an affine invariant sampler \cite{Shields2021a} were used to improve the robustness of the the subset simulation framework for highly nonlinear limit state functions and/or {high}-dimensional inputs. {There has also been interest in using machine learning models such as neural networks \cite{Papadopoulos2012a} and support vector regression \cite{Bourinet2011a} for replacing expensive {HF model} evaluations to compute the function $\widetilde{F}(\pmb{X})$}. 

Briefly, the subset simulation procedure entails the following. An intermediate failure probability value $p_o$ is first assigned (0.1 is typical). Monte Carlo is used to simulate $N$ samples of $\widetilde{F}(\pmb{X})$ in the first subset. This subset's failure threshold (i.e., $\mathcal{F}_1$) is set such that a fraction of the samples equal to $p_o$ exceed this threshold. MCMC is used to simulate $N$ samples of $\widetilde{F}(\pmb{X})$ in the second subset, {conditioned} upon these samples exceeding the threshold $\mathcal{F}_1$. As with the first subset, the second subset's failure threshold (i.e., $\mathcal{F}_2$) is set such that a fraction of the samples in this subset equal to $p_o$ exceeds this threshold. Subsequent subsets are similarly simulated using MCMC, until a significant number of samples exceed the required failure threshold $\mathcal{F}$. Equation \eqref{eqn:SS1} is used to estimate the failure probability, wherein the intermediate probabilities $P_1,\dots,P_{s|s-1},\dots,P_{N-1|N-2}$ are all equal to $p_o$, and the final conditional probability $P_{N|N-1}$ is equal to the fraction of samples exceeding the threshold $\mathcal{F}$. \citet{Au2001a,Au2014a} provide a more detailed description of this procedure.

{
\section{Active learning for high-fidelity versus low-fidelity model selection in subset simulation}\label{sec:HF_LF}}

{Here, we are interested in conducting subset simulation using multifidelity models. In particular, we have a high-fidelity (HF) and a low-fidelity (LF) model and here we devise an active learning strategy to determine when HF model calls are necessary (and when LF models calls are sufficient) within each conditional level of the subset simulation.}

In reference to Figure \ref{Schematic}, if the HF and LF models respectively take the random vectors $\pmb{X}_{HF}$ and $\pmb{X}_{LF}$ as inputs, whose outputs are defined as:

\begin{equation}
    \label{eqn:HFLF_1}
    \begin{aligned}
    y_{HF} &= F(\pmb{X}_{HF})~~~~~\textrm{High-fidelity model output}\\
    y_{LF} &= f(\pmb{X}_{LF})~~~~~~\textrm{Low-fidelity model output}\\
    \end{aligned}
\end{equation}

\noindent If $\pmb{X} = \pmb{X}_{HF} \cup \pmb{X}_{LF}$ is the superset of all the input parameters required by either the HF or LF model, then the required model output $\widetilde{F}(\pmb{X})$ for a given sample in the subset simulation is:

\begin{equation}
\label{eqn:HFLF_2}
\widetilde{F}(\pmb{X}) = \begin{cases}
F(\pmb{X}_{HF})~~~~~~~~~~~~~~\text{for HF model evaluation} \\
f(\pmb{X}_{LF})+\bar{\epsilon}(\pmb{X})~~~~~\text{for LF model evaluation} \\
\end{cases}
\end{equation}

\noindent where $\bar{\epsilon}(\pmb{X})$ is a mean correction term to the LF model prediction given by {a} $\mathcal{GP}$ {surrogate}. As such, the $\mathcal{GP}$ is initially trained to learn the differences {between} HF and LF model outputs, and it operates on the superset ${\pmb{X}}$. Given a new sample of input parameters, the $\mathcal{GP}$ correction term follows the posterior distribution defined in Equation \eqref{eqn:GP_posterior}. 

{
\subsection{Multi-fidelity active learning with standard Monte Carlo}

In reliability analysis, it is not necessary to know the true value of the performance function $\widetilde{F}(\cdot)$ at any given point. Rather, it is important only to correctly identify the sign of the performance function $G(\pmb{X})=\widetilde{F}(\pmb{X})-\mathcal{F}$ where negative values correspond to ``failure'' and positive values correspond to ``safe'' conditions. In a single-fidelity (HF) standard Monte Carlo setting, active learning has been used by several researchers \cite{bichon2008efficient, Echard2011a, sundar2019reliability} to determine when to make HF models calls and when to employ a $\mathcal{GP}$ surrogate model. One popular method, termed Adaptive-Kriging with Monte Carlo Simulation (AK-MCS) developed by \citet{Echard2011a} makes this determination by estimating the probability that the $\mathcal{GP}$ surrogate will incorrectly predict the sign of $G(\pmb{X})$. To estimate this probability, they developed the so-called $U$-function defined as follows:

\begin{equation}
    \label{eqn:u_0}
    U = \frac{|\mu_{\hat{G}}(\pmb{X})|}{\sigma_{\hat{G}}(\pmb{X})}
\end{equation}

\noindent where $\mu_{\hat{G}}(\pmb{X})$ is the mean $\mathcal{GP}$ prediction and $\sigma_{\hat{G}}(\pmb{X})$ is its standard deviation, such that $\Phi(-U)$ is the probability of incorrect sign prediction where $\Phi(
\cdot)$ is the standard normal CDF. Model evaluations are selected at points where ${U}$ is small, corresponding to areas where $\widetilde{F}(\pmb{X})$ is close to $\mathcal{F}$ and/or $\sigma_{\hat{G}}(\pmb{X})$ is large. Generally, sampling continues until the minimum $U$-value exceeds {a threshold $\mathcal{U}$; typically $\mathcal{U}=2$} which corresponds to the probability of making a sign error of $\Phi(-2)\approx0.0228$.}

{It is natural to extend this learning framework to a multi-fidelity modeling setting where, instead of evaluating the confidence in a surrogate model to predict the correct sign, we evaluate the confidence for a LF model with a $\mathcal{GP}$ correction to predict the correct sign. In this setting, our LF model is deemed sufficient when it has a high probability of accurately predicting the correct sign of $G(\pmb{X})$, after applying a $\mathcal{GP}$ correction term. In a standard Monte Carlo setting, the determination for} when an HF model call should be made in Equation \eqref{eqn:HFLF_2} depends on the probability of making a sign error (i.e., either a false positive or false negative characterization of failure) at the required failure threshold $\mathcal{F}$ when using the LF model. {To estimate this probability of making a sign error, the $U$-function can be adapted for multi-fidelity models as follows:}

\begin{equation}
    \label{eqn:u_1}
    U_{MF} = \frac{|f(\pmb{X}_{LF})+\bar{\epsilon}(\pmb{X})-\mathcal{F}|}{\sigma_{\epsilon}(\pmb{X})}
\end{equation}

\noindent where $\bar{\epsilon}(\pmb{X})$ is the mean $\mathcal{GP}$ {correction} and {$\sigma_{\epsilon}(\pmb{X})$} is its standard deviation. {As in the conventional AK-MCS, the multi-fidelity $U_{MF}$ assumes a small value when $f(\pmb{X}_{LF})+\bar{\epsilon}(\pmb{X})$ is close to $\mathcal{F}$ and/or when $\sigma_{\epsilon}(\pmb{X})$} is large. 

{This multi-fidelity AK-MCS (MF-AK-MCS) method, leveraging the multi-fidelity $U_{MF}$, provides a natural extension of the standard AK-MCS for multi-fidelity modeling.}


{
\subsection{Coupled active learning and subset simulation}

Huang et al.\ \cite{Huang2016a} showed that it can be beneficial to combine the standard AK-MCS framework described above with subset simulation in the AK-SS (Adaptive Kriging with Subset Simulation) method. In particular, they propose a two-step procedure that begins with AK-MCS and follows with a subset simulation using the established $\mathcal{GP}$ model. They then iterate with additional AK-MCS samples if the coefficient of variation of the $P_f$ estimate is too high. Again, it is natural to substitute a multi-fidelity model with a $\mathcal{GP}$ correction within this framework [thus using the $U_{MF}$ in Eq.\ \eqref{eqn:u_1} in place of the conventional $U$-function in Eq.\ \eqref{eqn:u_1}].} 

While the $U$-function presented in Equation \eqref{eqn:u_0} {[multi-fidelity $U$-function in Eq.\ \eqref{eqn:u_1}]} provides a means to check the performance of {a $\mathcal{GP}$ (LF model)} call near the required failure threshold $\mathcal{F}$, its robustness for {estimating} smaller failure probabilities ({on} the order $1\times10^{-5}$) {when combined with subset simulation} requires some discussion. {It has recently been shown \cite{Lelievre2018a, Razaaly2020a} that AK-MCS alone can break down for small failure probabilities due to the inability to sufficiently sample deep into the low-probability regions. This inability limits AK-MCS from identifying a sufficient number of candidates to add to the training set. The AK-SS method will also suffer from this drawback for low-failure probabilities due to the fact that the subset simulations and the adaptive Kriging are uncoupled. Moreover, the fact that subset simulation and adaptive Kriging are uncoupled means that, although the Kriging model will confidently predict the sign of the true performance function, it is not guaranteed to adequately model the intermediate limit surfaces. This means that the AK-SS estimate may have high variance due to incorrect estimates of intermediate conditional probabilities. At worst, the AK-SS may break down because the Kriging model is highly inaccurate in the intermediate subsets where little training data exists.}

{These issues can potentially be resolved by coupling the AK-MCS and subset simulation in the following way. Instead of conducting a conventional MCS to establish the training data, the training data are actively identified during subset simulation. More specifically, we draw a small number of samples from which to initially train the $\mathcal{GP}$ and then initiate subset simulation using this $\mathcal{GP}$ such that, for each new sample drawn in the subset simulation, we additionally evaluate the $U$-function. If {$U<\mathcal{U}$}, then a model evaluation is called and the $\mathcal{GP}$ is retrained. }

\begin{figure}[h]
\centering  
\includegraphics[width=2.5in, height=2.5in]{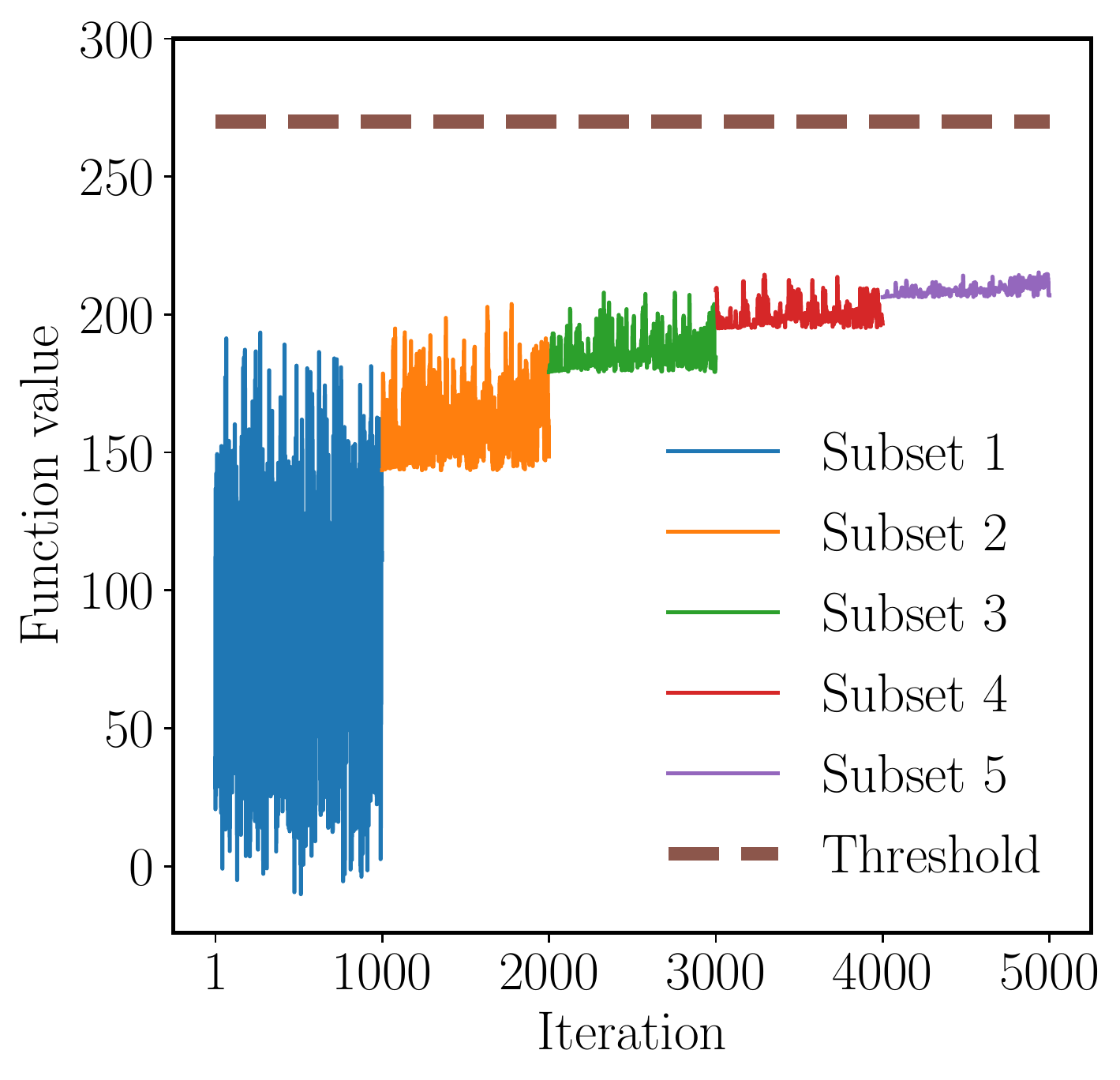} 
\caption{Breakdown of the active learning procedure under the traditional $U$-function. For the details on the function used, refer to Section \ref{sec:borehole}.}
\label{Failure_Case}
\end{figure}

{This method is robust and efficient when failure probabilities are modest, but still breaks down for very low failure probabilities. This is due to the fact that, for low failure probabilities, an insufficient number of subset simulation samples approaches the limit surface to force retraining of the $\mathcal{GP}$ (i.e. the subset simulation samples at every conditional level possess {$U>\mathcal{U}$}). This leaves the limit surface inadequately resolved in the $\mathcal{GP}$ and stalls the subset simulations as illustrated in Figure \ref{Failure_Case}. From this figure, we can see that the subset simulations never approach the true limit surface $\mathcal{F}=270$ due to a failure to retrain the under-resolved $\mathcal{GP}$. }


To overcome this problem, we {propose} a subset-dependent $U$-function, $U_s$, for the subsets $1\leq s < N_s$ defined as:
{
\begin{equation}
    \label{eqn:u_s}
    U_s = \frac{|\mu_{\tilde{F}}(\pmb{X})-\mathcal{F}_s|}{\sigma_{\tilde{F}}(\pmb{X})}~~~~\text{for }~1\leq s < N_s
\end{equation}
where $\mu_{\tilde{F}}(\pmb{X})$ and $\sigma_{\tilde{F}}(\pmb{X})$ are the mean and standard deviation of the $\mathcal{GP}$ surrogate for $\widetilde{F}(\pmb{X})$ and $\mathcal{F}_s$ is the threshold for conditional level $s$,}
estimated dynamically as sampling in subset $s$ progresses. For each new sample in this subset, $\mathcal{F}_{s}$ is computed as the $(1-p_o)^{\textrm{th}}$ quantile value of all the required predictions made by $\widetilde{F}(\pmb{X})$ thus far. Such a dynamically computed $\mathcal{F}_{s}$ quickly converges to the ``true'' failure threshold for this subset, due to the law of large numbers. 
{Finally, for} subset $s=N_s$, $U_s$ is defined as:

\begin{equation}
    \label{eqn:u_3ss}
    U_s = \frac{|\mu_{\tilde{F}}(\pmb{X})-\mathcal{F}|}{\sigma_{\tilde{F}}(\pmb{X})}~~~~\text{for }s=N_s
\end{equation}

\noindent where the check for the correct sign is made near the {true} failure threshold $\mathcal{F}$ in order to accurately trace the failure boundary and estimate $P_f$. 


{
\subsection{Coupled multi-fidelity active learning and subset simulation}
Merging the concepts of multi-fidelity AK-MCS and the coupled AK-SS presented in the previous sections results in a robust approach to adaptively select high-fidelity versus low-fidelity model evaluations within a subset simulation to estimate small failure probabilities. In particular, redefining our subset-dependent $U$-function to leverage a low-fidelity model with a $\mathcal{GP}$ correction, yields the following
\begin{equation}
    \label{eqn:u_3}
    U_{s}^{MF} = \frac{|f(\pmb{X}_{LF})+\bar{\epsilon}(\pmb{X})-\mathcal{F}_s|}{\sigma(\pmb{X})}~~~~\text{for }~1\leq s < N_s
\end{equation}
Using this subset-dependent, multi-fidelity learning function, we adaptively select to run the HF model during each conditional simulation when {$U_{s}^{MF}<\mathcal{U}$} and subsequently retrain the $\mathcal{GP}$ correction. Once again, for the final subset we have
\begin{equation}
    \label{eqn:u_3b}
    U_{s}^{MF} = \frac{|f(\pmb{X}_{LF})+\bar{\epsilon}(\pmb{X})-\mathcal{F}|}{\sigma(\pmb{X})}~~~~\text{for }s=N_s
\end{equation}
Details for implementation of this proposed multi-fidelity active learning approach are provided in the following section.
}

\section{Multi-fidelity active learning subset simulation}\label{sec:MF_SS_AL}

{In this section, we detail the step-by-step procedure for the proposed multi-fidelity active learning subset simulation method and derive coefficient of variation estimates for the resulting probabilities of failure.

\subsection{Proposed algorithm}

Subset simulation achieves variance reduction by expressing $P_f$ as a product of intermediate probabilities (i.e., $P_{s|s-1}$) and computing them individually. {To initialize the algorithm, we first decide upon the intermediate failure probability, $p_0$, and the number of simulations to draw in each conditional level $N$.}

\subsubsection{Initial training set}
The first step is to draw a small number ($N_{init}$) of Monte Carlo samples of $\pmb{X}=\pmb{X}_{LF} \cup \pmb{X}_{HF}$ from the distribution $q(\pmb{X})$ and evaluate both the LF model, $f(\pmb{X}_{LF})$, and the HF model,  $F(\pmb{X}_{HF})$. Next, evaluate the difference between these model evaluations as $\epsilon(\pmb{X})=F(\pmb{X}_{HF})-f(\pmb{X}_{LF})$ and train a $\mathcal{GP}$ surrogate for this discrepancy as $\hat{\epsilon}(\pmb{X})$ having mean $\bar{\epsilon}(\pmb{X})$ (the prediction) and standard deviation $\sigma(\pmb{X})$.}


{
\subsubsection{First conditional level}
}

{Since the first subset uses {standard} Monte Carlo, selection of the HF versus LF models is {relatively} straightforward. For each new sample of parameters $\pmb{X} = \pmb{X}_{HF} \cup \pmb{X}_{HF}$ drawn from the distribution $q(\pmb{X})$, a LF model evaluation is made and then corrected by the $\mathcal{GP}$, as presented in Equation \eqref{eqn:HFLF_2}. We then estimate the conditional failure threshold $\mathcal{F}_1$ by the $(1-p_0)$ quantile from the corrected LF model or HF model evaluations made thus far. For each corrected LF model evaluation, we evaluate $U_1^{MF}(\pmb{X})$ using Equation \eqref{eqn:u_3}. If {$U_1^{MF}(\pmb{X})<\mathcal{U}$}, we evaluate the HF model and then retrain the $\mathcal{GP}$ correction. This procedure is repeated sequentially for all $N$ samples during the Monte Carlo procedure. Algorithm \ref{alg:Proposed_1} further details this procedure.  } 

\begin{algorithm}
\small
   \caption{Active learning with multifidelity modeling (First conditional level)}
   \label{alg:Proposed_1}
    \begin{algorithmic}[1]
    \Require $\pmb{X} = \pmb{X}_{HF} \cup \pmb{X}_{LF}$, $q(\pmb{X})$, $F(\pmb{X}_{HF})$, $f(\pmb{X}_{LF})$, $\widetilde{F}(\pmb{X})$, $\mathcal{F}$, $N_s$, $N$, $N_{c}$, $N_{dim}$, $p_o$
    \State ${\epsilon}(\pmb{X}) = F(\pmb{X}_{HF}) - f(\pmb{X}_{LF}) = \mathcal{GP}\big(m(\pmb{X}), k(\pmb{X},\pmb{X}^\prime)\big)$ [can be further expanded using Equation \eqref{eqn:GP_posterior}]
    \For{$i = 1:N$}
    \State $\pmb{X}_i \sim q(\pmb{X})$
    \State $f_{i} = f(\pmb{X}_{LF,i})$
    \State Compute $U^{MF}_{1,i}$ using Equation \eqref{eqn:u_3} with $\mathcal{F}_1 = \mathbf{SORT}_{p_o}^{{\tilde{F}}}(\widetilde{F}_{1:i-1})$
    \If{$U^{MF}_{1,i} \geq \mathcal{U}$}
    \State $\widetilde{F}(\pmb{X}_i) = \widetilde{F}_i = f_{i} + \bar{\epsilon}_i$ {[accept LF model evaluation]}
    \Else
    \State $F_{i} = F(\pmb{X}_{HF,i})$ and $\widetilde{F}(\pmb{X}_i) = \widetilde{F}_i = F_{i}$ {[perform HF model evaluation]}
    \State $\epsilon(\pmb{X}_i) = \epsilon_i = F_i - f_i$
    \State $\epsilon = \big(\epsilon,~\epsilon_i\big)$ and $\pmb{X} = \big(\pmb{X},~\pmb{X}_i\big)$
    \State $\epsilon(\pmb{X}) = \mathcal{GP}\big(m(\pmb{X}), k(\pmb{X},\pmb{X}^\prime)\big)$~~~[$\mathcal{GP}$ re-training]
    \EndIf
    \EndFor
     \end{algorithmic}
\end{algorithm}
\normalsize


{
\subsubsection{Intermediate conditional levels}
}
For subsets $s>1$, MCMC is used to draw conditional samples and simulate the probability  {$P_{s|s-1}$. From the previous subset, we begin with a set of $p_0\times N$ samples lying within the new conditional level. From each of these conditional samples, we propagate a Markov chain}
using a component-wise Metropolis Hastings method {to generate new samples $\pmb{X}$ according to the conditional distribution}. 

{Prior to drawing MCMC samples, we first establish an initial estimate of the conditional failure threshold $\mathcal{F}_s$ as the $(1-p_0)$ quantile from the initial samples in the conditional level. We then initiate the MCMC algorithm by drawing a candidate sample, $\pmb{X}_j^*$, from the proposal distribution $p(\pmb{X}_j)$, centered around a previously accepted value $\pmb{X}_j$ for chain $j$ and computing the component-wise modified Metropolis-Hastings acceptance/rejection criterion (see \cite{Au2001a}). That is, for each component $X_{jk}^*$ of $\pmb{X}_j^*$, we evaluate

\begin{equation}
    \label{eqn:alg_1}
    \alpha_k = \frac{q(X_{jk}^*)~p(X_{jk})}{q(X_{jk})~p(X_{jk}^*)}
\end{equation}

\noindent where $p(X_{jk})$ is the marginal proposal density for dimension $k$ and $q(X_{jk})$ is the marginal density of the $k^{th}$ component of $\pmb{X}$, and accept the sample component with probability $\min\{1, \alpha_k
\}$. 

{For each accepted candidate, we evaluate the LF model and apply the $\mathcal{GP}$ correction. We then estimate the conditional failure threshold $\mathcal{F}_s$ by the $(1-p_0)$ quantile from the corrected LF model or HF model evaluations made thus far. Next, we evaluate the subset-dependent multi-fidelity $U_s^{MF}(\pmb{X}^*)$ from Eq.\ \eqref{eqn:u_3}. This $U$-function evaluation corresponds to a check of whether the LF model is sufficiently accurate for assessment of the failure criterion for conditional level $s$ (that is checking if the LF is sufficient to assess $f(\pmb{X}_{LF}^*)+\bar{\epsilon}(\pmb{X}^*)<\mathcal{F}_{s}$). If {$U_s^{MF}(\pmb{X}^*)<\mathcal{U}$}, then a HF model evaluation is called. For every HF model evaluation, the difference $F(\pmb{X}_HF)-f(\pmb{X}_{LF})$ is computed and the $\mathcal{GP}$ surrogate correction is retrained.

Next, we check whether that accepted sample lies in the conditional level $s$. If the model output $\widetilde{F}(\pmb{X})$ (i.e., either the corrected LF output or the HF output) is greater than $\mathcal{F}_{s-1}$, we accept the sample. Otherwise, we reject it. This process proceeds until $N$ samples are drawn from conditional level $s$. Generally speaking, most of these samples will correspond to LF model evaluations and a small number of new HF model evaluations will be introduced in the vicinity of conditional threshold $\mathcal{F}_s$. This process is repeated for each conditional level, with each conditional failure probability equal to $p_0$, until the final one. Algorithm \ref{alg:Proposed_2} further details this procedure.}}



{
\subsubsection{Final conditional level}
In the final conditional level, we reach a state where $\mathcal{F}_s>\mathcal{F}$ and we therefore replace the intermediate failure condition ($\mathcal{F}_s$) with the true failure condition
($\mathcal{F}$) in each of the steps of the previous section. More specfically, the multi-fidelity $U_s^{MF}$ from Eq.\ \eqref{eqn:u_3b} is used to identify when to evaluate the HF model.

Finally, we can estimate the conditional failure probability for the final subset as
\begin{equation}
    p_{N_s} = \dfrac{N_{\tilde{F}\geq \mathcal{F}}}{N}
\end{equation}
where $N_{\tilde{F}\geq \mathcal{F}}$ is the number of failure samples in the final conditional level. Using conventional subset simulation estimators, the probability of failure can ultimately be computed as:
\begin{equation}
    P_f = P_1\prod_{i=2}^{N_s}P_{s|s-1}=p_0^{N_s-1}p_{N_s}
\end{equation}
However, since our multi-fidelity model has some probability of incorrect sign prediction (albeit small when properly trained), an improved probability of failure estimate can be devised that accounts for this potential error. This is derived next.
}



\begin{algorithm}
\small
   \caption{Active learning with multifidelity modeling (intermediate and final conditional levels)}
   \label{alg:Proposed_2}
    \begin{algorithmic}[1]
    \For{$s = 2:N_s$}
    \State $\pmb{S_{\widetilde{F}}} = \mathbf{SORT}_{p_o}^{{\tilde{F}}}\big(\widetilde{\pmb{F}}^{s-1}\big)$ and $\pmb{S} = \mathbf{SORT}_{p_o}^{{\tilde{F}}}\big(\widetilde{\pmb{X}}^{s-1}\big)$ and $\widetilde{F}_{lim} = \mathbf{MIN}\big(\pmb{S_{\widetilde{F}}}\big)$
    \For{$i = 1:N_{c}$}
    \State $\widetilde{{F}}^{s}_{i,1} = \pmb{S}_{\widetilde{F},i}$ and $\pmb{X}^s_{i,1} = \pmb{S}_i$
    \For{$k = 1:\mathbf{int}(N/N_{c})$}
    \For{$j = 1:N_{dim}$}
    \State {sample} ${X}^*_{j} \sim p({X}_{i,k-1,j})$
    \State $\ln{\alpha} = \ln{q({X}^*_{j})} + \ln{p({X}_{i,k-1,j})} - \ln{q({X}_{i,k-1,j})} - \ln{p({X}^*_{j})}$
    \If{$\ln{\alpha} \geq \ln{\mathbf{RAND}}$}
    \State ${X}_{i,k,j} = {X}^*_{j}$
    \Else
    \State ${X}_{i,k,j} = {X}_{i,k-1,j}$
    \EndIf
    \EndFor
    \State $f_{i,k} = f(\pmb{X}_{LF,i,k})$
    \State Compute $U^{MF}_{s,ik}$ using either Equation \eqref{eqn:u_3} (with $\mathcal{F}_s = \mathbf{SORT}_{p_o}^{{\tilde{F}}}(\widetilde{F}_{1:i,1:k-1})$) or \eqref{eqn:u_3b}
    \If{$U^{MF}_{s,ik} \geq \mathcal{U}$}
    \State $\widetilde{F}^*  = f_{i,k} + \bar{\epsilon}_{i,k}$ {[accept LF model evaluation]} 
    \Else
    \State $F_{i,k} = F(\pmb{X}_{HF,i,k})$ and $\widetilde{F}^* = F_{i,k}$ {[perform HF model evaluation]}
    \State $\epsilon(\pmb{X}_{i,k}) = \epsilon_{i,k} = F_{i,k} - f_{i,k}$
    \State $\epsilon = \big(\epsilon,~\epsilon_{i,k}\big)$ and $\pmb{X} = \big(\pmb{X},~\pmb{X}_{i,k}\big)$
    \State $\epsilon(\pmb{X}) = \mathcal{GP}\big(m(\pmb{X}), k(\pmb{X},\pmb{X}^\prime)\big)$~~~[$\mathcal{GP}$ re-training]
    \EndIf
    \If{$\widetilde{F}^* \geq \widetilde{F}_{lim}$}
    \State $\widetilde{F}(\pmb{X}^s_{i,k}) = \widetilde{F}_{i,k} = \widetilde{F}^*$
    \Else
    \State $\widetilde{F}(\pmb{X}_{i,k}) = \widetilde{F}_{i,k} = \widetilde{F}_{i,k-1}$
    \EndIf
    \EndFor
    \EndFor
    \EndFor
     \end{algorithmic}
\end{algorithm}
\normalsize

\subsection{Estimators for the intermediate failure probabilities and corresponding coefficients of variation}\label{cov_proposed}

For the first subset, which relies on Monte Carlo sampling, an estimator for the intermediate failure probability $P_1$ is given by:

\begin{equation}
\label{eqn:mean_1}
\begin{aligned}
P_1 \approx \hat{P}_1 &= \frac{1}{N}~\sum_{i=1}^N \mathcal{P}_i\\
&= \frac{1}{N}~\sum_{i=1}^N P(\mathbf{I}_{i}=1 | \mathbf{I}_{i, {{LF}}} = 1)~P( \mathbf{I}_{i, {{LF}}} = 1) + P(\mathbf{I}_{i}=1 | \mathbf{I}_{i, {{LF}}} = 0)~P( \mathbf{I}_{i, {{LF}}} = 0)\\
\end{aligned}
\end{equation}

\noindent where $\mathbf{I}_{{i}}$ is an indicator function for an output value {$i$} exceeding the first subset's failure threshold $(\mathcal{F}_1)$ and $\mathbf{I}_{{{i}}, {{LF}}}$ is an indicator function for the $\mathcal{GP}$ {corrected LF model} predicting the correct sign at the first subset's failure threshold. $\mathcal{P}_i$ is the probability that $\mathbf{I}_{{{i}}}=1$, { and is expanded according to the law of total probability in Eq.\ \eqref{eqn:mean_1}. Note that, in general $\hat{P}_1\ne p_0$, although it is straightforward to show that $\hat{P}_1=p_0$ when a perfect LF model is used.}
$\mathcal{P}_i$ in Equation \eqref{eqn:mean_1} can be further written as:

\begin{equation}
\label{eqn:cov_99}
\mathcal{P}_{{i}} = P(\mathbf{I}_{{{i}}}=1) = \begin{cases}
1 \times {\Phi}_{{{i}}} + 0 \times (1-{\Phi}_{{{i}}}) = {\Phi}_{{{i}}}~~~~~~~~~~~~~~~\text{if $\mathbf{I}_{{{i}},LF} = 1$} \\
0 \times {\Phi}_{{{i}}} + 1 \times (1-{\Phi}_{{{i}}})  = 1-{\Phi}_{{{i}}}~~~~~~~~~~\text{if $\mathbf{I}_{{{i}},LF} = 0$} 
\end{cases}
\end{equation}

\noindent where ${\Phi}_{{{i}}}$ is the probability of the $\mathcal{GP}$ {corrected LF model evaluated at point $i$} predicting the correct sign. Since {} the $\mathcal{GP}$ prediction follows a normal distribution and {$U_1^{MF}$} computed using Equation \eqref{eqn:u_3} is a standard normal random variable, {${\Phi}_{{i}}=\Phi(-U_{1,i}^{MF})$ is evaluated  using the standard normal cdf}. An estimator for the coefficient of variation (COV; $\gamma_1$) of $\mathcal{P}_1$ is {subsequently} given by:
\begin{equation}
\label{eqn:cov_98}
\gamma_1 \approx \hat{\gamma}_1 = \sqrt{\frac{1-\hat{P}_1}{\hat{P}_1~N}}
\end{equation}
For subsequent subsets (i.e., $s>1$) reliant on MCMC sampling, an estimator for the conditional failure probability is given by:
\begin{equation}
\label{eqn:mean_2}
P_{s|s-1} \approx \hat{P}_{s|s-1} = \frac{1}{N}~\sum_{i=1}^{N_{c}}~\sum_{k=1}^{N/N_{c}} \mathcal{P}^s_{ik}
\end{equation}

\noindent where, for $1<s\leq N_s$, $\mathcal{P}_{ik}^s$ is {again expanded according to the law of total probability as}:

\begin{equation}
\label{eqn:cov_2}
\mathcal{P}_{ik}^s = P(\mathbf{I}_{ik}^s=1) = P(\mathbf{I}_{ik}^s=1 | \mathbf{I}_{ik, {LF}}^{s} = 1)~P( \mathbf{I}_{ik, {LF}}^{s} = 1) + P(\mathbf{I}_{ik}^s=1 | \mathbf{I}_{ik, {LF}}^{s} = 0)~P( \mathbf{I}_{ik, {LF}}^{s} = 0)
\end{equation}

\noindent where $\mathbf{I}_{ik}^s$ is an indicator function for the output value {$(i,k)$} exceeding the $s^{\textrm{th}}$ subset's failure threshold ($\mathcal{F}_s$), and $\mathbf{I}_{ik, {LF}}^{s}$ is an indicator function for the $\mathcal{GP}$ {corrected LF model} predicting the correct sign at the $s^{\textrm{th}}$ subset's failure threshold, given the $k^{\textrm{th}}$ sample in the ${i}^{\textrm{th}}$ Markov chain. Equation \eqref{eqn:cov_2} can be simplified to:

\begin{equation}
\label{eqn:cov_3}
\forall~1<s\leq N_s,~\mathcal{P}_{ik}^s = \begin{cases}
1 \times {\Phi}_{ik}^{s} + 0 \times (1-{\Phi}_{ik}^{s}) = {\Phi}_{ik}^{s}~~~~~~~~~~~~~~~\text{if $\mathbf{I}_{ik{,LF}}^s = 1$} \\
0 \times {\Phi}_{ik}^{s} + 1 \times (1-{\Phi}_{ik}^{s})  = 1-{\Phi}_{ik}^{s}~~~~~~~~~~\text{if $\mathbf{I}_{ik{,LF}}^s = 0$} 
\end{cases}
\end{equation}

\noindent where ${\Phi}_{ik}^{s} = P( I_{ik, {,LF}}^{s} = 1){=\Phi(U_{s,ik}^{MF})}$ is the probability of $\mathcal{GP}$ {corrected LF model }predicting the correct sign at the $s$ subset threshold {$\mathcal{F}_s$}. 
The probability $P(I_{ik}^s=1|.)$ in Equation \eqref{eqn:cov_2} equals the value of the indicator function $I_{ik}^s$ itself. 
${\Phi}_{ik}^{s}$ {again denotes the} standard normal cumulative distribution function.
The variance in the conditional failure probability estimator{, following from \cite{Au2001a}}, is given by:

\begin{equation}
\label{eqn:cov_1}
\mathbf{E}(\hat{P}_{s|s-1} - P_{s|s-1})^2 = \frac{1}{N^2}~\sum_{i=1}^{N_c}\mathbf{E}\bigg[\sum_{k=1}^{N/N_c}\mathcal{P}_{ik}^s-P_{s|s-1}\bigg]^2
\end{equation}

\noindent Following Au and Beck (2001) \cite{Au2001a}, in Equation \eqref{eqn:cov_1}:

\begin{equation}
\label{eqn:cov_4}
\mathbf{E}\bigg[\sum_{k=1}^{N/N_c}\mathcal{P}_{ik}^s-P_{s|s-1}\bigg]^2 = \sum_{k,l=1}^{N/N_c} \mathbf{E}\big[(\mathcal{P}_{il}^s-P_{s|s-1})~(\mathcal{P}_{il+k}^s-P_{s|s-1})\big] = \sum_{k,l=1}^{N/N_c} R_s(k-l)
\end{equation}

\noindent $R_s(k-l)$ in this equation can be expanded as:

\begin{equation}
\label{eqn:cov_5}
\begin{aligned}
R_s(k-l) &= \mathbf{E}(\mathcal{P}^s_{il}~\mathcal{P}^s_{il+k}) - \mathbf{E}(\mathcal{P}^s_{il})~P_{s|s-1} - \mathbf{E}(\mathcal{P}^s_{il+k})~P_{s|s-1} + P_{s|s-1}^2\\
&\approx \mathbf{E}(\mathcal{P}^s_{il}~\mathcal{P}^s_{il+k}) - P_{s|s-1}^2 \equiv \hat{R}_s(k-l)\\
\end{aligned}
\end{equation}

\noindent Again, following \citet{Au2001a}, the variance estimator for the $s^\textrm{th}$ subset failure probability can be expressed as:

\begin{equation}
\label{eqn:cov_6}
\hat{\sigma}_s^2=\mathbf{E}(\hat{P}_{s|s-1} - P_{s|s-1})^2 = \frac{1}{N}~\bigg[\hat{R}_s(0)+2~\sum_{k=1}^{N/N_c-1}\bigg(1-\frac{kN_c}{N}~\hat{R}_s(k)\bigg)\bigg]
\end{equation}

\noindent where $\hat{R}_s(0) = \mathbf{Var}(\mathcal{P}_{ik}^s) = P_{s|s-1}~(1-P_{s|s-1})$. Equation \eqref{eqn:cov_6} can be further expressed as:

\begin{equation}
\label{eqn:cov_7}
\hat{\sigma}_s^2 = \frac{\hat{P}_{s|s-1}~(1-\hat{P}_{s|s-1})}{N}~(1+\hat{\gamma}_s)~~~~\text{where,}~\hat{\gamma}_s = 2 \sum_{k=1}^{N/N_c-1}\bigg(1-\frac{kN_c}{N}~\hat{\rho}_s(k)\bigg)
\end{equation}

\noindent and $\hat{\rho}_s(k) = \hat{R}_s(k)/\hat{R}_s(0)$ is the autocorrelation coefficient at lag $k \in \{1,\dots,N/N_c-1\}$. The COV for subset $s$ $(1<s \leq N_s)$ is given by:

\begin{equation}
\label{eqn:cov_8}
\hat{\delta}_s = \sqrt{\frac{1-\hat{P}_{s|s-1}}{N~\hat{P}_{s|s-1}}~(1+\hat{\gamma}_s)}
\end{equation}

\noindent As the autocorrelation coefficient $\hat{\rho}_s(k) \to 0$, $\hat{\gamma}_s \to 0$ in Equation \eqref{eqn:cov_8}, the COV estimator for subset $s$ converges to that of a Monte Carlo COV estimator. However, for practical applications, the MCMC samples can be correlated and $\hat{\rho}_s(k)>0$, indicating that the COV estimator would be greater than that of a Monte Carlo COV estimator. {Note also that correlation between chains can be further included using the extension derived in \cite{Shields2021a}}. The total COV estimator, considering all the subsets, is given by:

\begin{equation}
\label{eqn:cov_9}
\hat{\delta} = \sqrt{\sum_{s=1}^{N_s}\hat{\delta}_s^2}~~\forall~~1\leq s \leq N_s
\end{equation}

\noindent A comparison between the COV estimators from Equation \eqref{eqn:cov_9} and that proposed by \citet{Au2001a}, which does not consider the use of a $\mathcal{GP}$ for modeling fidelity selection, is presented in Section \ref{sec:fourb}.

\section{Description of the case studies}

The proposed framework for active learning with multifidelity modeling is demonstrated using two sets of case studies: (1) standard academic case studies and (2) FE model case studies. Standard academic case studies {use a mathematical function as the HF model} and enable us to easily compare the proposed algorithm's performance to that of a direct Monte Carlo method. They also enable a visualization of the proposed algorithm's capability to trace the failure boundaries for low-dimensional input parameters. Since there is flexibility in the choice of LF model in the proposed algorithm, either a $\mathcal{GP}$ or a DNN trained with a few evaluations of the HF model is used for these academic cases. Three standard academic case studies are considered: (1a) the four-branch function is a simple, low-dimensional function that permits visualization of the failure boundary; (1b) the Rastrigin function is a complex, low-dimensional function that permits visualization of the failure boundaries; and (1c) the Borehole function is a higher-dimensional function used to compare the performance of a $\mathcal{GP}$ and a DNN as the LF model.

FE model case studies can involve a time-consuming HF model (treated to be ``exact'') and a faster-running LF model that may ignore some of the HF model characteristics (e.g., physics, model parameters and mesh complexity). These case studies enable us to evaluate the scalability of the proposed algorithm for more realistic applications. Two FE model case studies are considered: (2a) four-sided lid-driven cavity with the steady-state Navier-Stokes as the HF model and the steady-state Stokes approximation (i.e., the nonlinear convective term is ignored) as the LF model; and (2b) computation of the maximum von Mises stress in a 3-D domain with a finely-meshed transversely isotropic material as the HF model and a coarsely-meshed isotropic material as the LF model. It is noted that, in Case Study (2b), the HF and LF models do not share the same number of input parameters. Additionally, since the HF model for the FE case studies is computationally expensive to run under Monte Carlo, the standard subset simulation is used as a reference to evaluate the performance of the proposed algorithm. Table \ref{Table_Cases} summarizes the case studies considered in this paper.

\begin{table}[h]
\centering
\caption{Description of the test cases for evaluating the performance of the proposed algorithm.}
\label{Table_Cases}
\small
\begin{tabular}{ |c|c|c|c|c|c| }
\hline
\textbf{No.} & \textbf{Case study} & \textbf{HF model} & \textbf{LF model} & \textbf{\# parameters} & \textbf{Notes}\\
\hline
\multicolumn{6}{|c|}{\textbf{Standard academic case studies}}\\
\hline
1a & Four-branch function & The function & $\mathcal{GP}$ prediction & 2 & \Centerstack[c]{Vizualizing the algorithm \\ performance under a \\ simple failure function}\\
\hline
1b & Rastrigin function & The function & $\mathcal{GP}$ prediction & 2 & \Centerstack[c]{Vizualizing the algorithm \\ performance under a \\ complex failure function}\\
\hline
1c & Borehole function & The function & \Centerstack[c]{$\mathcal{GP}$ prediction\\ or \\ DNN prediction} & 8 & \Centerstack[c]{Comparison between $\mathcal{GP}$ \\ and DNN as the \\ LF model} \\
\hline
\multicolumn{6}{|c|}{\textbf{Finite element case studies}}\\
\hline
2a & \Centerstack[c]{Four-sided \\ lid-driven cavity} & \Centerstack[c]{Navier-Stokes \\ equations} & \Centerstack[c]{Stokes \\ equations} & 6 & \Centerstack[c]{Ignored physics \\ in the LF model} \\
\hline
2b & \Centerstack[c]{Maximum von Mises \\ stress in a 3-D \\ cylindrical domain} & \Centerstack[c]{Transversely \\ isotropic  \\ material} & \Centerstack[c]{Isotropic material\\Coarser mesh} & \Centerstack[c]{8 (HF) \\ 5 (LF)} & \Centerstack[c]{Ignored material properties \\ and coarser mesh \\ in the LF model} \\
\hline
\end{tabular}
\begin{tablenotes}
\small
\item {\textbf{Abbreviations.} HF: High Fidelity; LF: Low Fidelity; DNN: Deep Neural Network; $\mathcal{GP}$: Gaussian process.}
\end{tablenotes}
\end{table}

\section{Standard academic case studies}\label{sec:academic}

In this section, we apply the proposed framework for active learning with multifidelity modeling to academic case studies and evaluate the framework's performance.

\subsection{Four-branch limit state function}\label{sec:fourb}

\begin{figure}[h]
\begin{subfigure}{0.5\textwidth}
\centering  
\includegraphics[width=3.0in, height=2.25in]{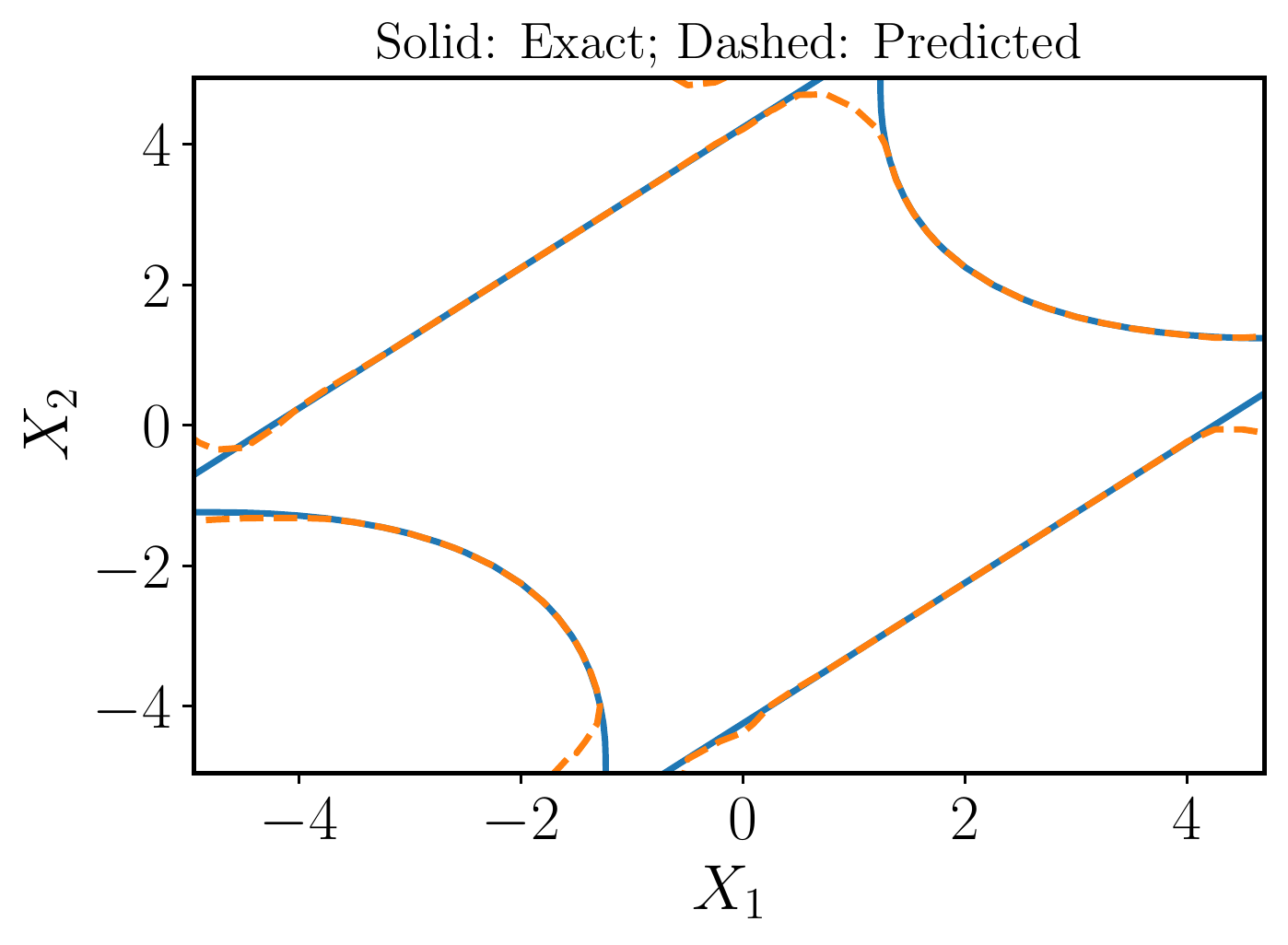} 
\caption{}
\label{FourBranch_1}
\end{subfigure}
\begin{subfigure}{0.5\textwidth}
\centering
\includegraphics[width=3.0in, height=2.25in]{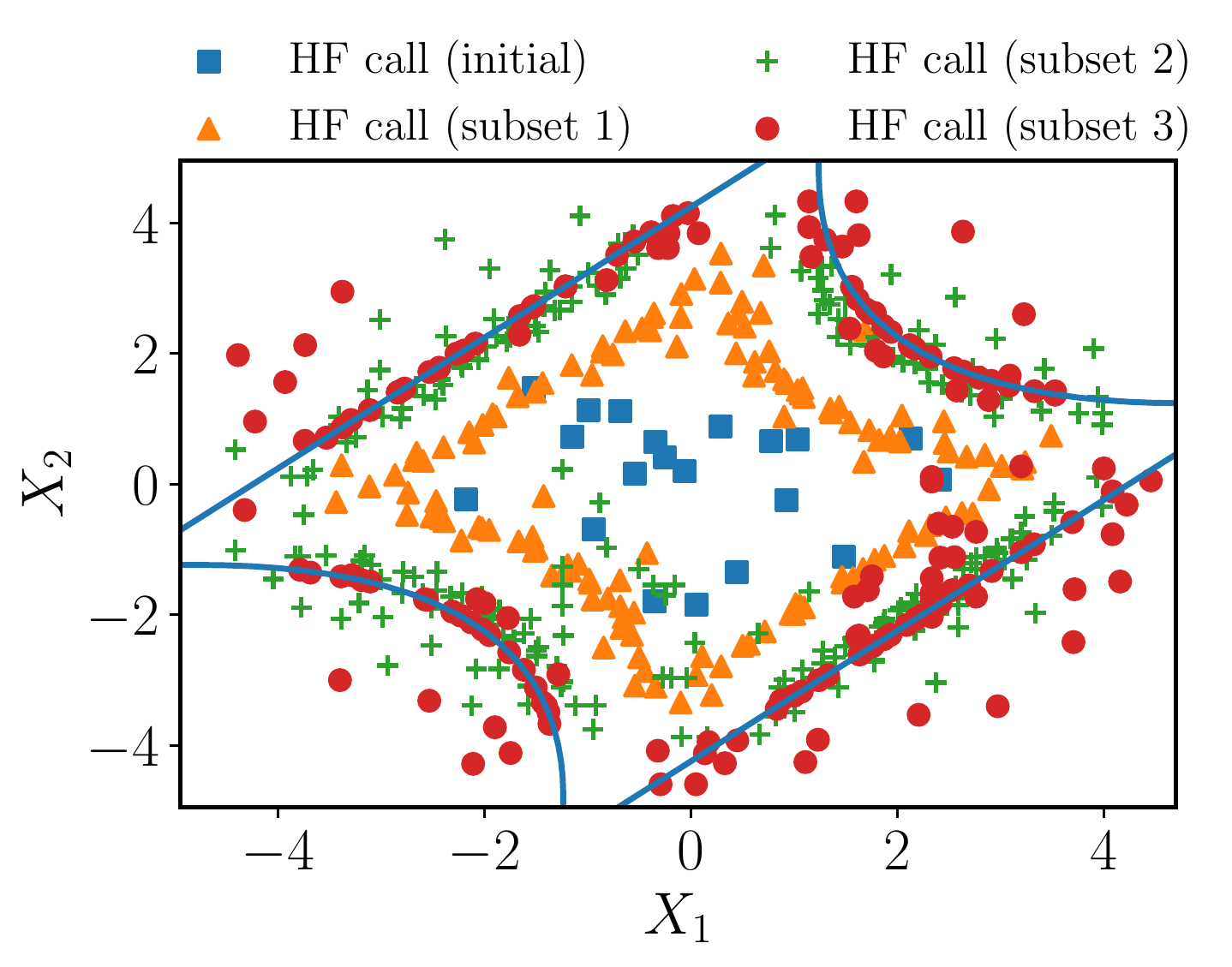} 
\caption{}
\label{FourBranch_2}
\end{subfigure}
\caption{(a) Comparison between the exact failure boundary and the failure boundary predicted by the {$\mathcal{GP}$ corrected LF model} in the proposed algorithm {at the end of the simulations}. (b) The exact failure boundary and the locations of the HF model calls across the three subsets.}
\label{FourBranch}
\end{figure}

The four-branch function is given by:

\begin{equation}
    \label{fb_1}
    F(\pmb{X}) = \mathbf{min}\begin{cases}
    3+(X_1-X_2)^2/10-(X_1+X_2)/\sqrt{2}\\
    3+(X_1-X_2)^2/10+(X_1+X_2)/\sqrt{2}\\
    (X_1-X_2)+6/\sqrt{2}\\
    (X_2-X_1)+6/\sqrt{2}\\
    \end{cases}
\end{equation}

\noindent where $\pmb{X} = \{X_1,~X_2\}$ are the two input parameters that follow a standard normal distribution. The failure threshold is $\mathcal{F}=0$. Equation \eqref{fb_1} is treated as the HF model. In the proposed algorithm, there is flexibility over the choice of LF model. A $\mathcal{GP}$ trained using 20 evaluations of the HF model is treated as the LF model. In the proposed algorithm, {our active learning} $\mathcal{GP}$ learns the differences between the HF and LF models. This $\mathcal{GP}$ is trained using 20 {different} evaluations of both the HF and LF models. With three subsets and 20,000 calls per subset of either the HF or LF model, the proposed algorithm is used {to estimate} $P_f$. Figure \ref{FourBranch_1} presents the contour of the exact failure boundary as well as the failure boundary predicted by the {$\mathcal{GP}$ corrected LF model at the end of all simulations}. It is noted that the exact and predicted failure boundary contours look mostly similar, except at the four corners {where {fewer HF} samples {are} available}. However, this mismatch near the boundaries can be rectified by {increasing the number of samples in each subset}. Figure \ref{FourBranch_2} presents the exact failure boundary with the locations of the HF model calls across the three subsets. For the first and second subsets, the HF calls are concentrated near the intermediate failure thresholds. Additionally, this threshold for the second subset is very close to the required failure threshold $\mathcal{F}=0$. For the third subset, the HF calls are concentrated near $\mathcal{F}=0$.

\begin{figure}[h]
\begin{subfigure}{0.5\textwidth}
\centering  
\includegraphics[width=2.4in, height=2.2in]{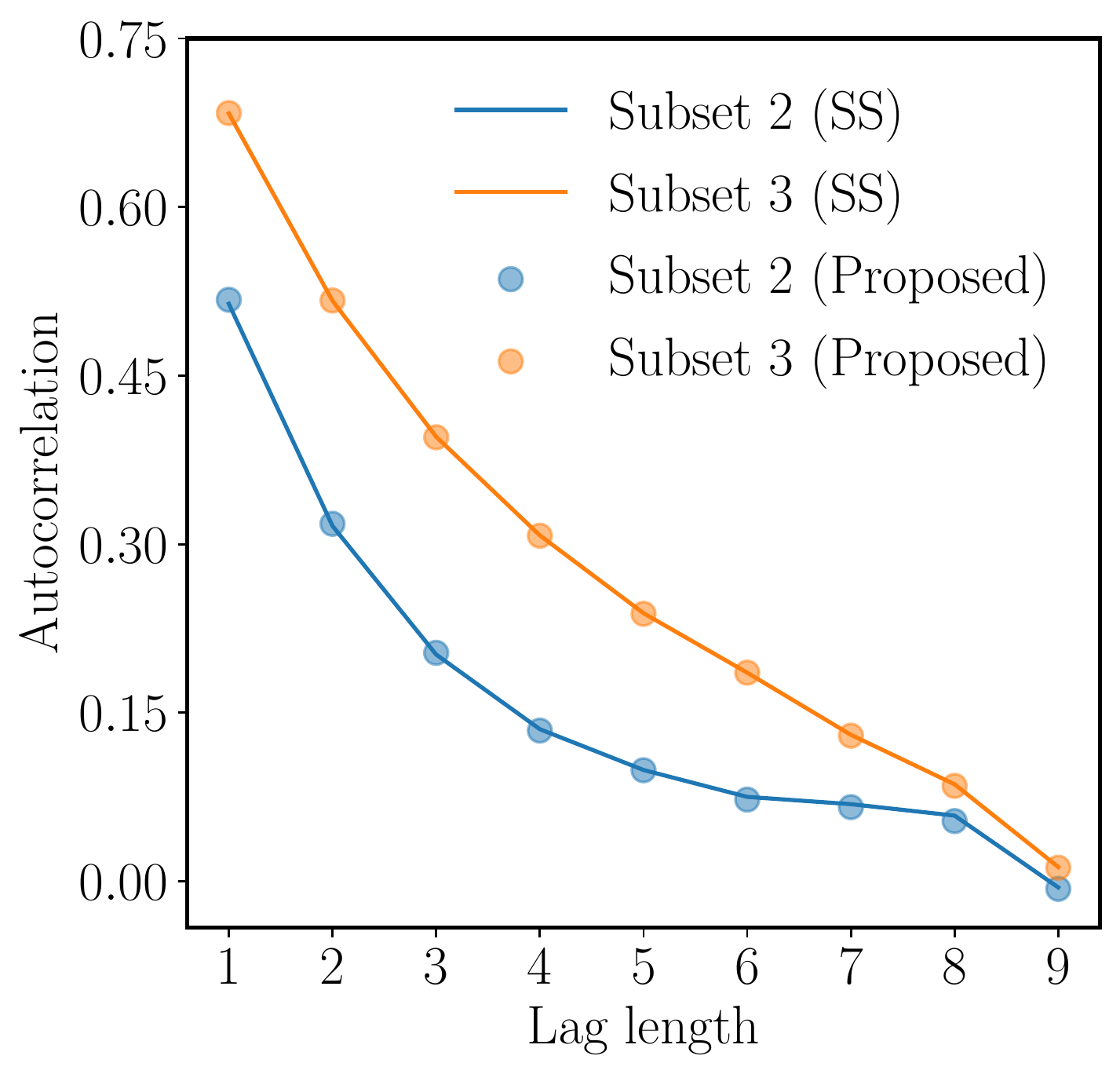} 
\caption{}
\label{FourBranch_diag_1}
\end{subfigure}
\begin{subfigure}{0.5\textwidth}
\centering
\includegraphics[width=2.4in, height=2.2in]{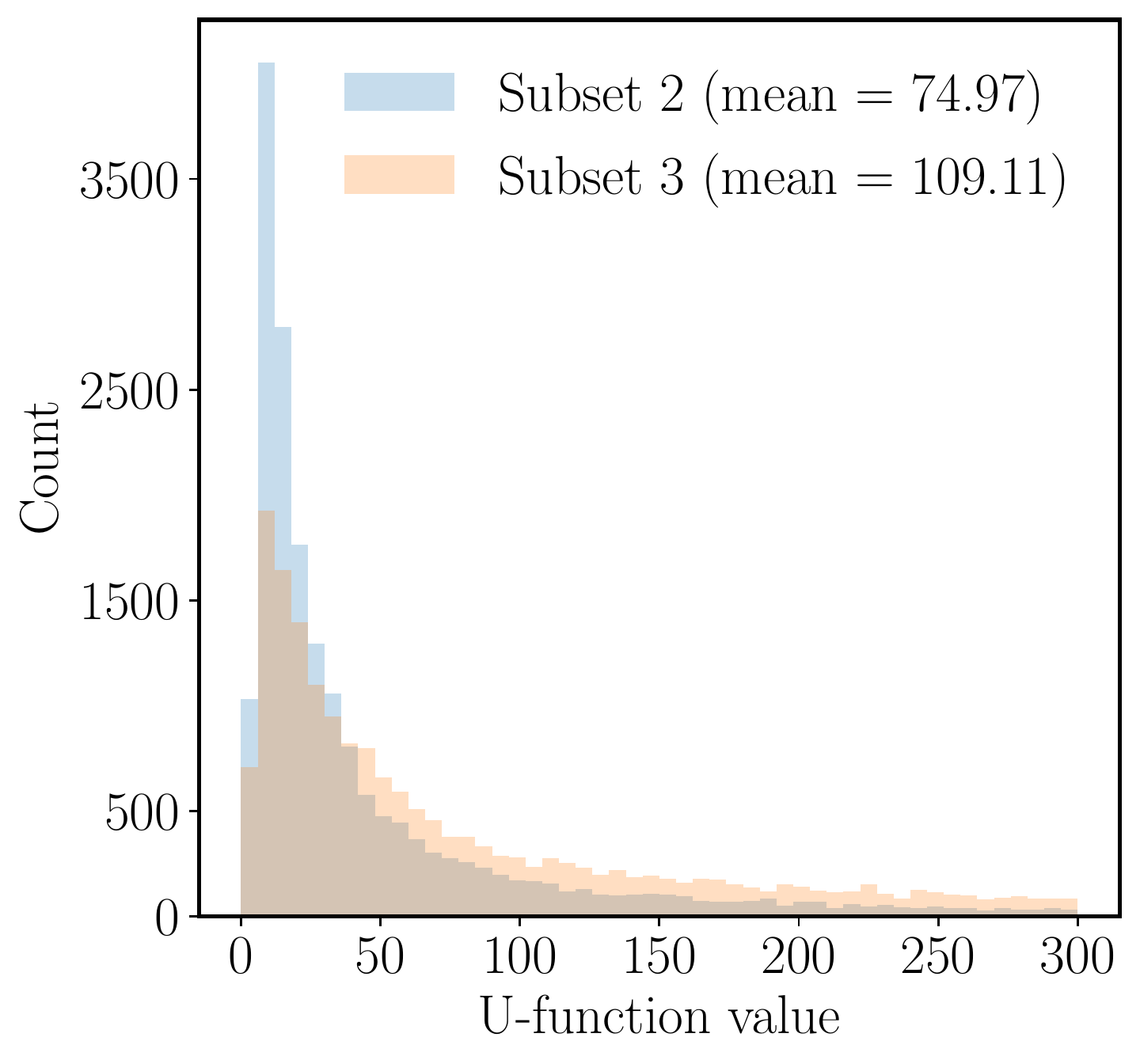} 
\caption{}
\label{FourBranch_diag_2}
\end{subfigure}
\caption{(a) Comparison of the autocorrelations between MCMC samples from subset simulation (SS) and the proposed algorithm. (b) Distribution of the $U$-function values in the proposed algorithm for Subsets 2 and 3.}
\label{FourBranch_diag}
\end{figure}

In Section \ref{cov_proposed}, an estimator for the COV using the proposed algorithm was suggested. This COV will now be discussed in comparison to the COV for subset simulation proposed by \citet{Au2001a}. These COVs differ in terms of how the autocorrelation term $\hat{\rho}_s(k) = \hat{R}_s(k)/\hat{R}_s(0)$ in Equation \eqref{eqn:cov_7} is defined. While subset simulation uses indicator functions in the autocorrelation term to characterize failures in a subset, the proposed algorithm, which relies on a $\mathcal{GP}$, uses probabilities (i.e., ${\mathcal{P}_i}$ and $\mathcal{P}_{ik}^s$ in Equations \eqref{eqn:mean_1} and \eqref{eqn:mean_2}, respectively). Therefore, a comparison between the autocorrelations of the proposed algorithm and the subset simulation gives an indication of {the differences in their} COVs. Figure \ref{FourBranch_diag_1} presents these autocorrelations for Subsets 2 and 3. While Subset 2 uses Equation \eqref{eqn:u_3} to compute the U-function, Subset 3, being the final subset, uses Equation \eqref{eqn:u_3b}. For both the subsets, it is noted that the proposed algorithm and the {original} subset simulation have near-identical autocorrelations, {and hence will have comparable COVs}. This match can be further examined through the $U$-function values for these two subsets, presented in Figure \ref{FourBranch_diag_2}. It is noted that most $U$-function values for both the subsets are substantially greater than the threshold $\mathcal{U}=2$, indicating a negligible probability of the $\mathcal{GP}$ making an error in selecting the HF versus LF model, as discussed in Section \ref{sec:MF_SS_AL}. {S}uch a negligible error means that the probabilities $\mathcal{P}_i$ and $\mathcal{P}_{ik}^s$ in Equations \eqref{eqn:mean_1}--\eqref{eqn:cov_5} converge to the indicator functions used in the COV formulation for subset simulation proposed by \citet{Au2001a}.

Table \ref{Table:FourBranch} compares Monte Carlo simulation, subset simulation, and {the} proposed algorithm in regard to $P_f$, COV, and number of HF calls. Results corresponding to two versions of the proposed algorithm are presented that use either subset-dependent $U$-functions (i.e., Equations \eqref{eqn:u_3} and \eqref{eqn:u_3b}) or a subset-independent $U$-function (i.e., Equation \eqref{eqn:u_1}). In all four cases, when a similar COV is applied, the $P_f$ values are {in agreement}. More importantly, the two versions of the proposed algorithm require only a fraction of the calls to the HF model, as compared to either Monte Carlo or subset simulation. Though using the subset-independent $U$-function requires fewer calls to the HF model than using the subset-dependent one, the latter is more robust under smaller $P_f$ values. In the present case, the $P_f$ is not small enough to show an advantage of using a subset-dependent $U$-function. Section \ref{sec:borehole} will discuss this further.

\begin{table}[h]
\centering
\caption{Results comparison among Monte Carlo, subset simulation, and the proposed algorithm, with both subset-dependent and -independent $U$-functions for the four-branch limit state function.}
\label{Table:FourBranch}
\small
\begin{tabular}{ |c|c|c|c|c| }
\hline
& \textbf{Monte Carlo} & \Centerstack[c]{\textbf{Subset}\\\textbf{simulation}$^{\dagger}$} & \Centerstack[c]{\textbf{Proposed algorithm}\\\textbf{with subset-dependent}\\\textbf{$U$-functions}$^{\dagger}$} & \Centerstack[c]{\textbf{Proposed algorithm}\\\textbf{with subset-independent}\\\textbf{$U$-function}$^{\dagger}$}\\ 
\hline
$\mathbf{P_f}$ & 4.32E-3 & 4.37E-3 & 4.46E-3 & 4.35E-3 \\ 
\hline
\textbf{COV} & 0.045 & 0.047 (0.031$^{\ddagger}$) & 0.045 (0.031$^{\ddagger}$) & 0.047 (0.031$^{\ddagger}$)\\ 
\hline
\textbf{$\#$ HF calls} & 110000 & 60000 & 490 & 247\\ 
\hline
\end{tabular}
\begin{tablenotes}
\item \normalsize{$^{\dagger}$} \scriptsize{Uses three subsets, with 20,000 samples for each}
\item \normalsize{$^{\ddagger}$} \scriptsize{COV value without considering the cross-correlations in the MCMC samples}
\end{tablenotes}
\end{table}

\subsection{Rastrigin limit state function}\label{sec:rastrigin}

\begin{figure}[h]
\begin{subfigure}{0.5\textwidth}
\centering  
\includegraphics[width=3.0in, height=2.25in]{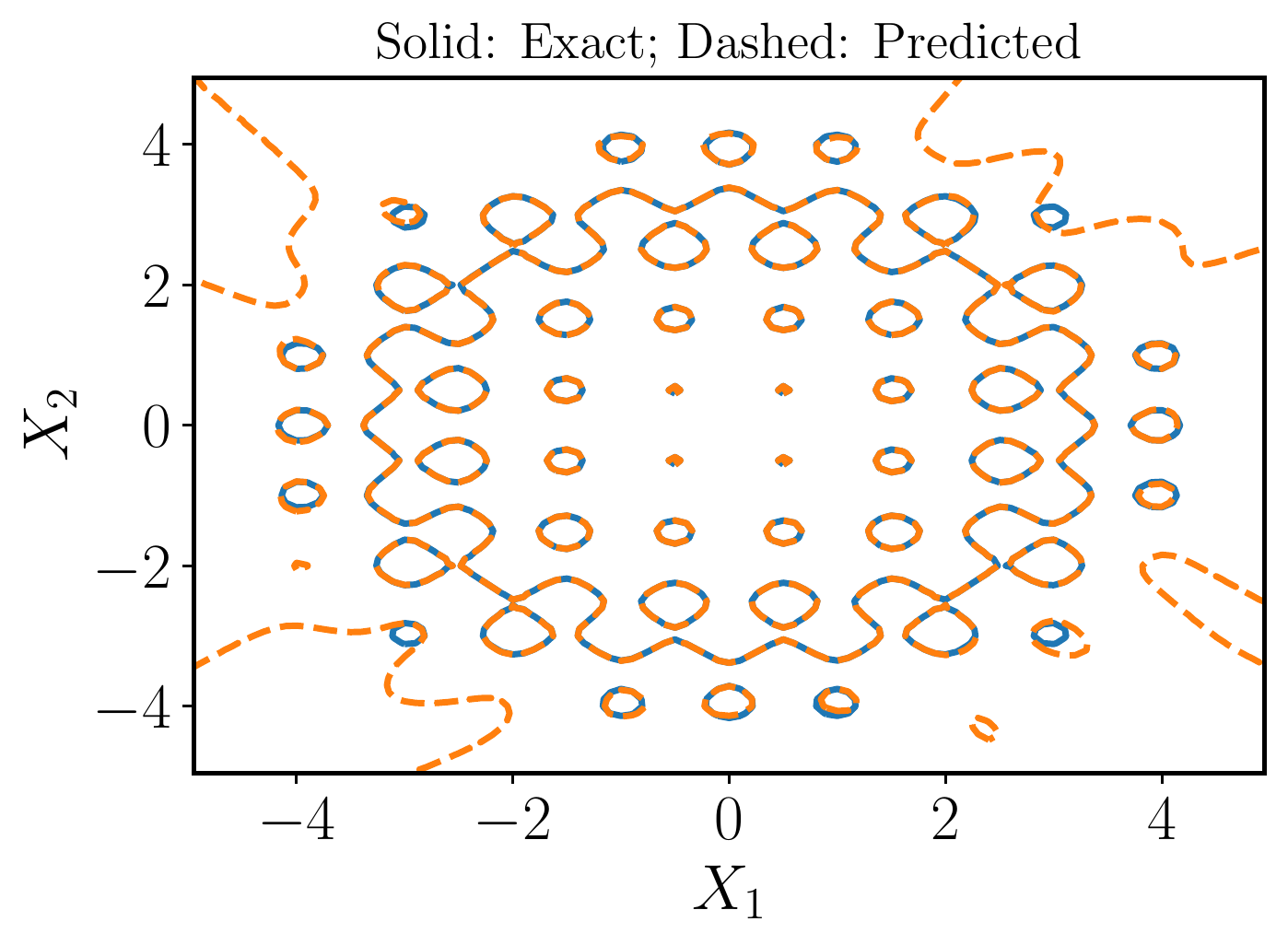} 
\caption{}
\label{Rastrigin_1}
\end{subfigure}
\begin{subfigure}{0.5\textwidth}
\centering
\includegraphics[width=3.0in, height=2.25in]{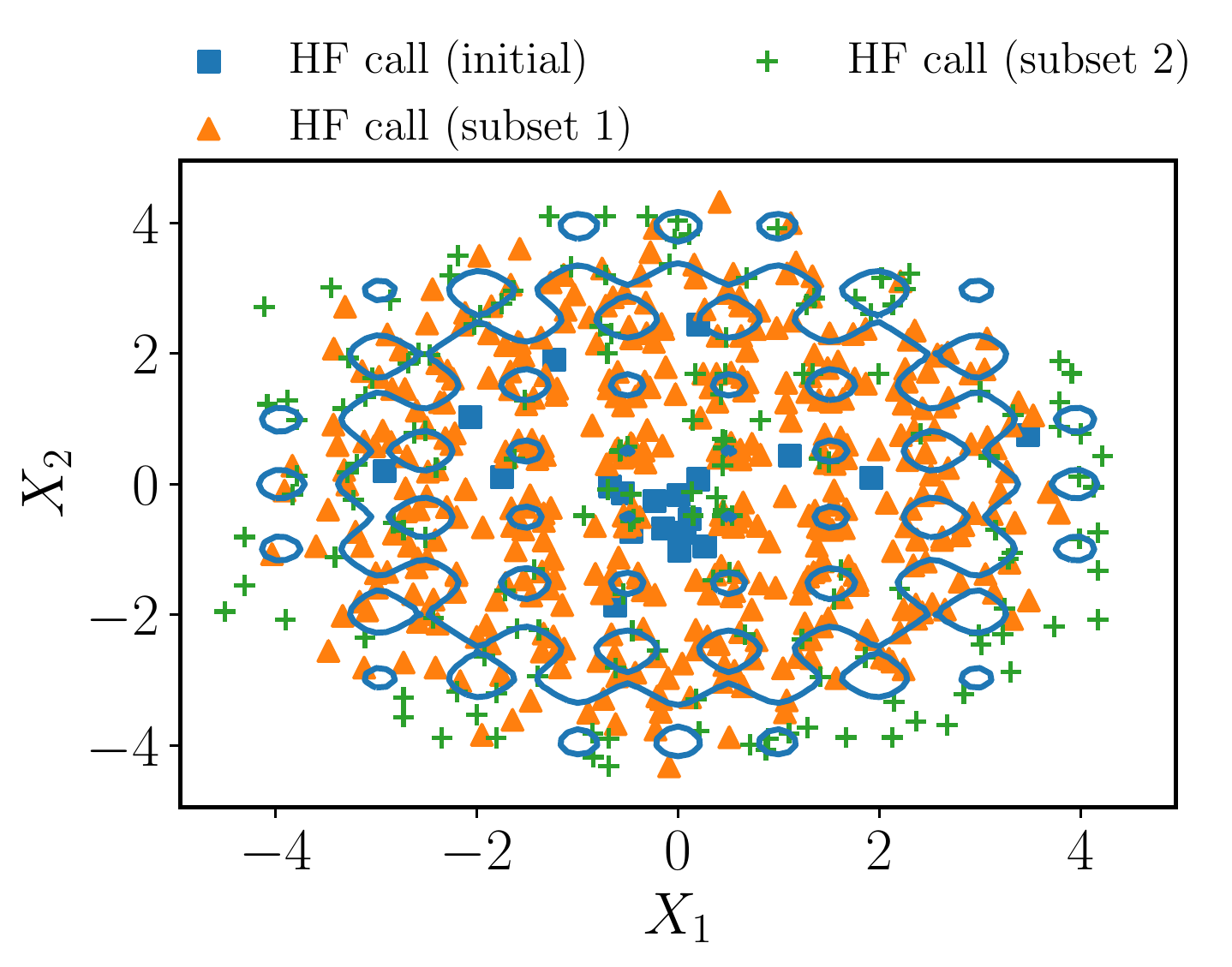} 
\caption{}
\label{Rastrigin_2}
\end{subfigure}
\caption{(a) Comparison of the exact failure boundary and the failure boundary predicted by the {final $\mathcal{GP}$ corrected LF model} in the proposed algorithm. (b) The exact failure boundary and the locations of the HF model calls across the two subsets.}
\label{Rastrigin}
\end{figure}

The Rastrigin function has a complex failure domain, and is given by:

\begin{equation}
    \label{eqn:Rastrigin_1}
    F(\pmb{X}) = 10-\sum_{i=1}^2 \big(X_i^2-5~\cos{(2\pi X_i)}\big)
\end{equation}

\noindent where $\pmb{X} = \{X_1,~X_2\}$ are the two input parameters that follow a standard normal distribution. The failure threshold is $\mathcal{F}=0$. Again, Equation \eqref{eqn:Rastrigin_1} is treated as the HF model, and $\mathcal{GP}$, trained with 20 evaluations of the HF model, is treated as the LF model. In the proposed algorithm, 20 {different} evaluations of the HF and LF models are used to {initially} train the {actively learning} $\mathcal{GP}$ to learn the differences between these models. With two subsets and 40,000 calls per subset of either the HF or LF model, the proposed algorithm is used {to estimate $P_f$.} Figure \ref{Rastrigin_1} presents the exact failure boundary, as well as the one predicted by the {final $\mathcal{GP}$-corrected LF model} in the proposed algorithm. Both of these failure boundaries look very similar, except near the edges where a smaller number of {HF} samples are typically available. Figure \ref{Rastrigin_2} presents the exact failure boundary with the locations of the HF model calls across the two subsets. It is noted that these HF calls are mostly concentrated around the failure boundary, though this may be less discernable in this example, given the complexity of the failure boundary. 

Table \ref{Table:Rastrigin} presents the $P_f$, COV, and number of calls to the HF model using Monte Carlo {simulation}, subset simulation, and the proposed algorithm. Across all three methods, the $P_f$ values are {in close agreement} when a similar COV is applied, although the proposed algorithm requires a fraction of calls to the HF model compared with either Monte Carlo or subset simulation. Additionally, the proposed algorithm with a subset-independent $U$-function requires fewer calls to the HF model than when using a subset-dependent $U$-function. The $P_f$ value in the present case is not small enough to notice the advantage of using a subset-dependent $U$-function, which offers more robustness when estimating smaller $P_f$ values.

\begin{table}[h]
\centering
\caption{Comparison of the results from Monte Carlo, subset simulation, and the proposed algorithm, with both subset-dependent and -independent $U$-functions for the Rastrigin limit state function.}
\label{Table:Rastrigin}
\small
\begin{tabular}{ |c|c|c|c|c| }
\hline
& \textbf{Monte Carlo} & \Centerstack[c]{\textbf{Subset}\\\textbf{simulation}$^{\dagger}$} & \Centerstack[c]{\textbf{Proposed algorithm}\\\textbf{with subset-dependent}\\\textbf{$U$-function}$^{\dagger}$} & \Centerstack[c]{\textbf{Proposed algorithm}\\\textbf{with subset-independent}\\\textbf{$U$-function}$^{\dagger}$}\\ 
\hline
$\mathbf{P_f}$ & 7.28E-2 & 7.23E-2 & 7.37E-2 & 7.28E-2\\ 
\hline
\textbf{COV} & 0.015 & 0.016 (0.015$^{\ddagger}$) & 0.016 (0.015$^{\ddagger}$) & 0.017 (0.015$^{\ddagger}$)\\ 
\hline
\textbf{$\#$ HF calls} & 60000 & 80000 & 724 & 581\\ 
\hline
\end{tabular}
\begin{tablenotes}
\item \normalsize{$^{\dagger}$} \scriptsize{Uses two subsets, with 40,000 samples for each}
\item \normalsize{$^{\ddagger}$} \scriptsize{COV value without considering the cross-correlations in the MCMC samples}
\end{tablenotes}
\end{table}
\normalsize

\subsection{Borehole limit state function}\label{sec:borehole}

The borehole function is given by:

\begin{equation}
    \label{Borehole_1}
    F(\pmb{X}) = \frac{2\pi~T_u~(H_u-H_l)}{\ln{(r/r_w)}~\bigg(1+\frac{2LT_u}{\ln{(r/r_w)}~r_w^2~K_w}+\frac{T_u}{T_l}\bigg)}
\end{equation}

\noindent where $F(\pmb{X})$ is the water flow and $\pmb{X} = \{r_w,~r,~T_u,~H_u,~T_l,~H_l,~L,~K_w\}$ is the input parameter {vector with parameters described in Table \ref{Table:Borehole_Inputs}} The failure threshold is $\mathcal{F} = 270$. Equation \eqref{Borehole_1} is treated as the HF model. The proposed algorithm is run independently using two LF models: (1) a $\mathcal{GP}$; or (2) a DNN with six neurons in the first hidden layer and four neurons in the second. Both the LF models are trained using 20 evaluations of the HF model. {The {active learning $\mathcal{GP}$} in the proposed algorithm is {initially} trained using 20 evaluations of the HF and LF models, to learn the differences in their predicted values.} With five subsets and 40,000 calls per subset of either the HF or LF model, the proposed algorithm is used {to estimate $P_f$} with subset-dependent $U$-functions (i.e., Equations \eqref{eqn:u_3} and \eqref{eqn:u_3b}). Table \ref{Table:Borehole} presents the results computed using Monte Carlo {simulation}, subset simulation, and the proposed algorithm, with either the $\mathcal{GP}$ or DNN as the LF model. For similar COV values, it is noted that the $P_f$ values across the different methods are not only very small, but also {in very good agreement with one another}. Using either a $\mathcal{GP}$ or DNN as the LF model, the proposed algorithm requires only a fraction of the calls to the HF model, as compared to either Monte Carlo or subset simulation. Additionally, there may be some advantage in using a DNN as the LF model, as it requires $16 \%$ fewer calls to the HF model as compared with using $\mathcal{GP}$ as the LF model. {This implies that, in this case, the DNN appears to provide a better LF model from the 20 training samples.} 

\begin{figure}[h]
\begin{subfigure}{0.5\textwidth}
\centering  
\includegraphics[width=2.8in, height=2.8in]{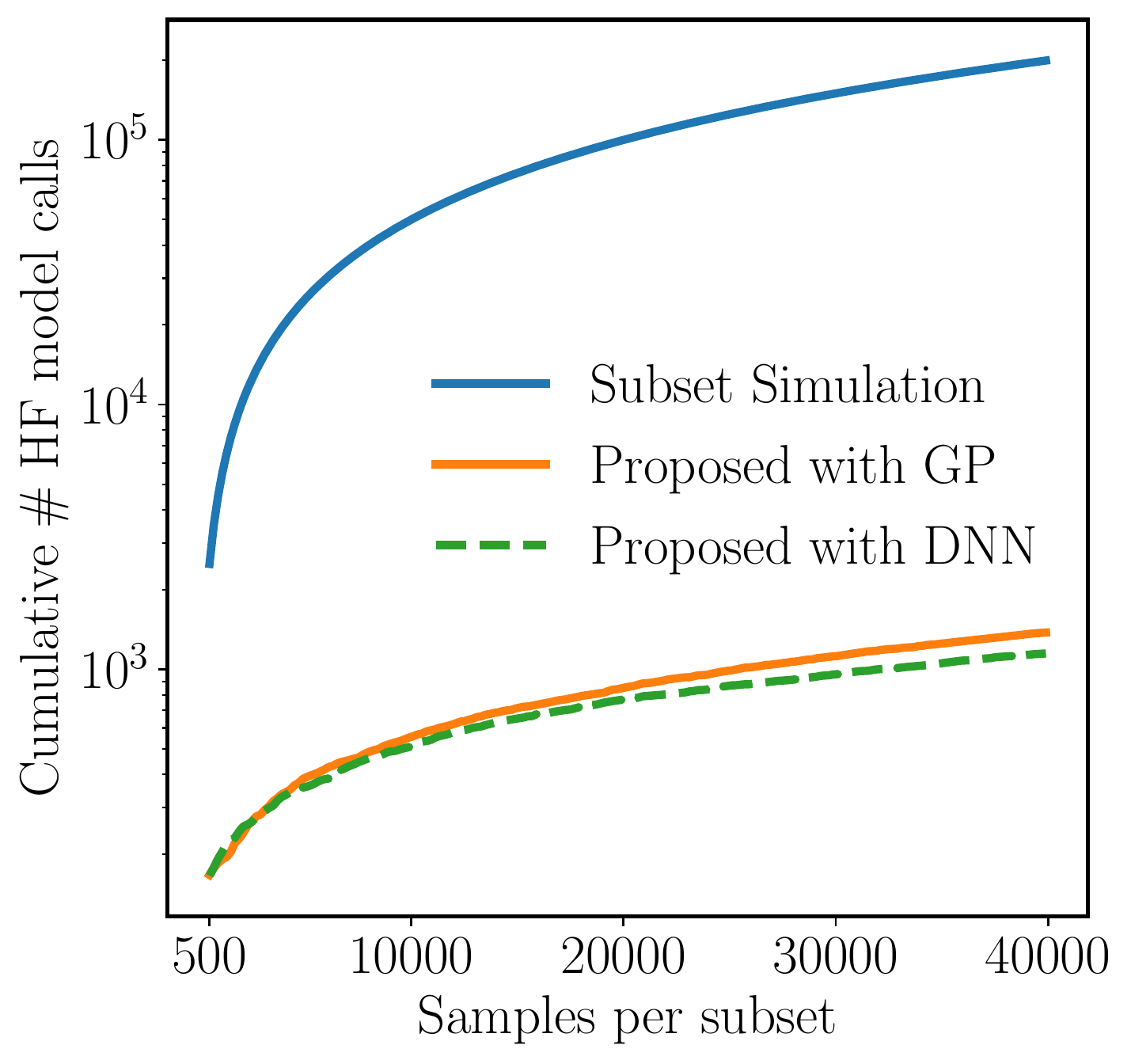} 
\caption{}
\label{Borehole_convergence_1}
\end{subfigure}
\begin{subfigure}{0.5\textwidth}
\centering
\includegraphics[width=2.9in, height=2.7in]{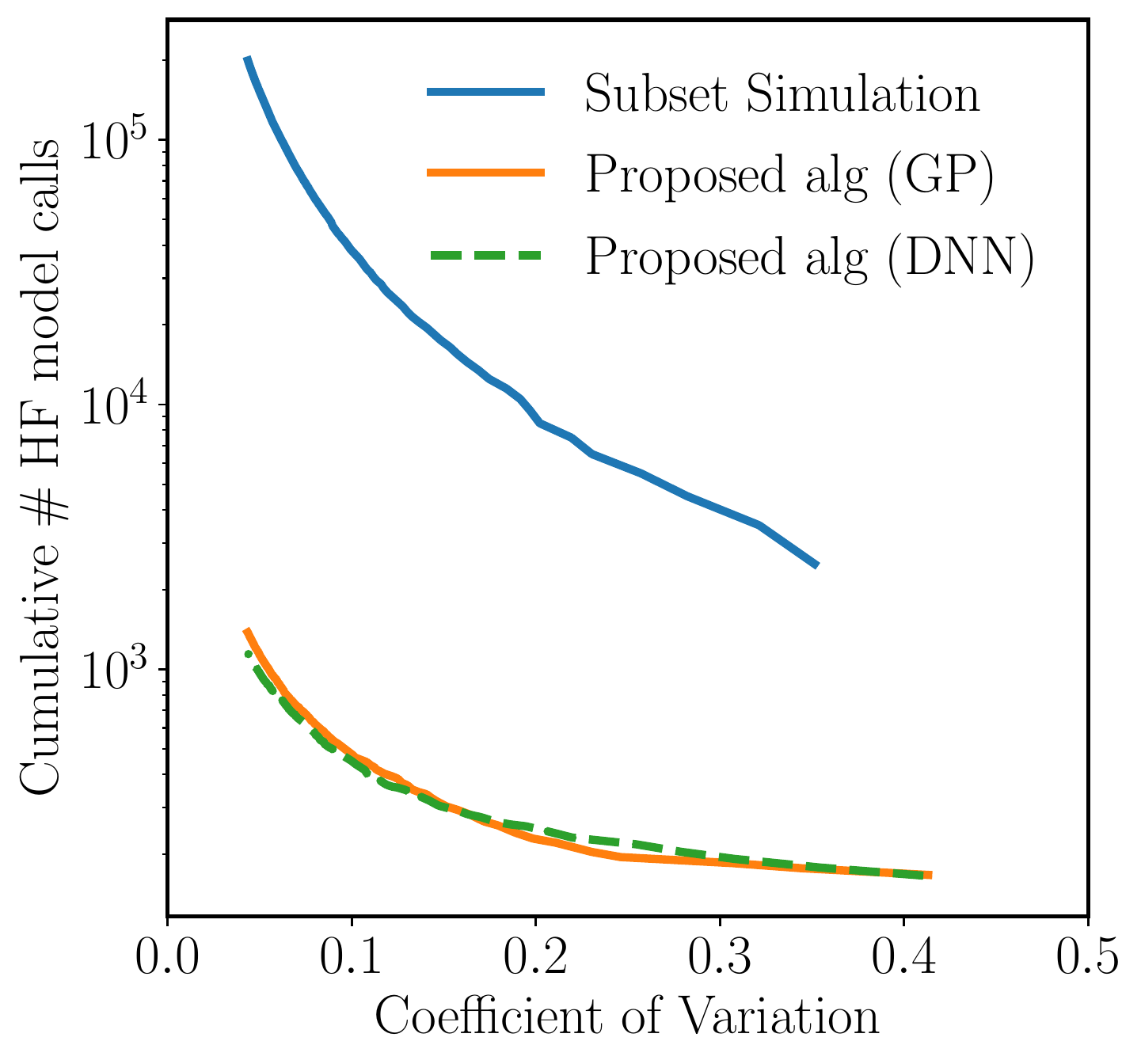} 
\caption{}
\label{Borehole_convergence_2}
\end{subfigure}
\caption{Cumulative number of calls to the high-fidelity model with (a) the number of samples per subset and (b) the coefficient of variation for the borehole limit state function.}
\label{Borehole_convergence}
\end{figure}

{To illustrate the scaling of the proposed algorithm,} Figure \ref{Borehole_convergence_1} presents the cumulative number of HF model calls across all subsets, with respect to the number of samples in each subset, for the three-subset-based methods. Figure \ref{Borehole_convergence_2} {further} presents the cumulative number of HF model calls with the {COV}.

\begin{table}[h]
\centering
\caption{Parameters of the borehole limit state function and their probability distributions.}
\label{Table:Borehole_Inputs}
\small
\begin{tabular}{ |c|c|c|c| }
\hline
\textbf{Variable} & \textbf{Definition} &  \textbf{Distribution} & \textbf{Parameters}\\ 
\hline
$r_w$ & Borehole radius & Uniform & $[0.05,~0.1]$ \\ 
\hline
$\ln{r}$ & Radius of influence & Normal & $[7.71,~1.0056]$ \\ 
\hline
$T_u$ & Upper aquifer transmissivity & Uniform & $[63070,~115600]$ \\ 
\hline
$H_u$ & Upper aquifer potentiometric head & Uniform & $[990,~1110]$ \\ 
\hline
$T_l$ & Lower aquifer transmissivity & Uniform & $[63.1,~116]$ \\
\hline
$H_l$ & Lower aquifer potentiometric head & Uniform & $[700,~820]$ \\ 
\hline
$L$ & Borehole length & Uniform & $[1120,~1680]$ \\ 
\hline
$K_w$ & Hydraulic conductivity & Uniform & $[9855,~12045]$ \\
\hline
\end{tabular}
\end{table}
\normalsize

\begin{table}[h]
\centering
\caption{Comparison of the results from Monte Carlo, subset simulation, and the proposed algorithm {for} the borehole limit state function$^{\diamond}$.}
\label{Table:Borehole}
\small
\begin{tabular}{ |c|c|c|c|c| }
\hline
& \textbf{Monte Carlo} & \Centerstack[c]{\textbf{Subset}\\\textbf{simulation}$^{\dagger}$} & \Centerstack[c]{\textbf{Proposed algorithm}\\\textbf{with $\mathcal{GP}$ as LF model}$^{\dagger}$} & \Centerstack[c]{\textbf{Proposed algorithm}\\\textbf{with DNN as LF model}$^{\dagger}$}\\ 
\hline
$\mathbf{P_f}$ & 2.83E-5 & 2.94E-5  & 2.92E-5 & 2.9E-5\\ 
\hline
\textbf{COV} & 0.045 &  0.043 (0.031$^{\ddagger}$) & 0.043 (0.031$^{\ddagger}$) &  0.043 (0.031$^{\ddagger}$)\\ 
\hline
\textbf{$\#$ HF calls} & 17,000,000 & 200,000 & 1379 & 1147\\ 
\hline
\end{tabular}
\begin{tablenotes}
\item \normalsize{$^{\diamond}$} \scriptsize{Subset-dependent $U$-functions are used in the proposed algorithm. The subset-independent $U$-function gives {$P_f=0$}}
\item \normalsize{$^{\dagger}$} \scriptsize{Uses five subsets, with 40,000 samples for each}
\item \normalsize{$^{\ddagger}$} \scriptsize{COV value without considering the cross-correlations in the MCMC samples}
\end{tablenotes}
\end{table}
\normalsize

\begin{figure}[h]
\begin{subfigure}{0.333\textwidth}
\centering  
\includegraphics[width=2in, height=2in]{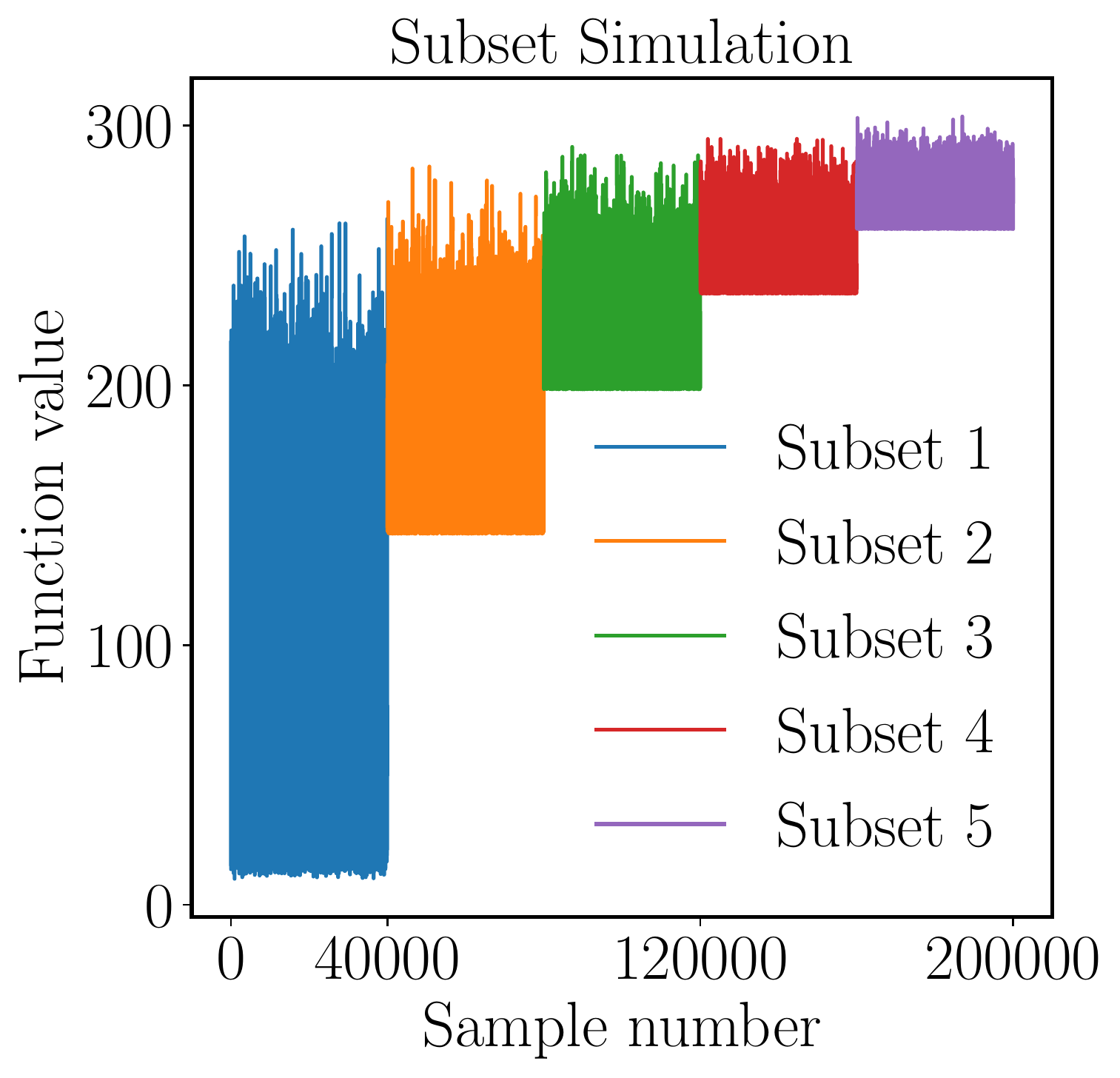} 
\caption{}
\label{Borehole_trace_1}
\end{subfigure}
\begin{subfigure}{0.333\textwidth}
\centering  
\includegraphics[width=2in, height=2in]{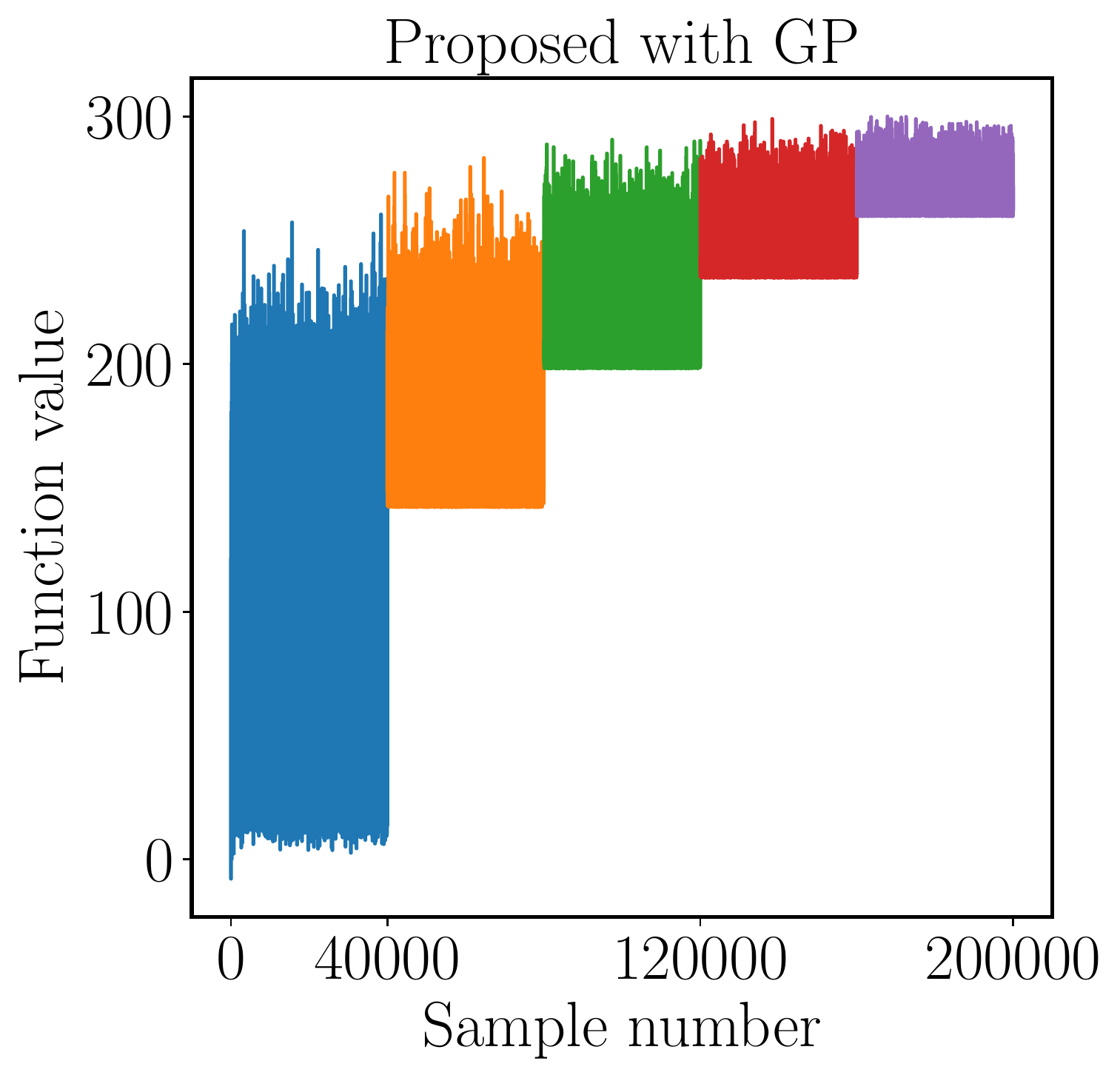} 
\caption{}
\label{Borehole_trace_2}
\end{subfigure}
\begin{subfigure}{0.333\textwidth}
\centering  
\includegraphics[width=2in, height=2in]{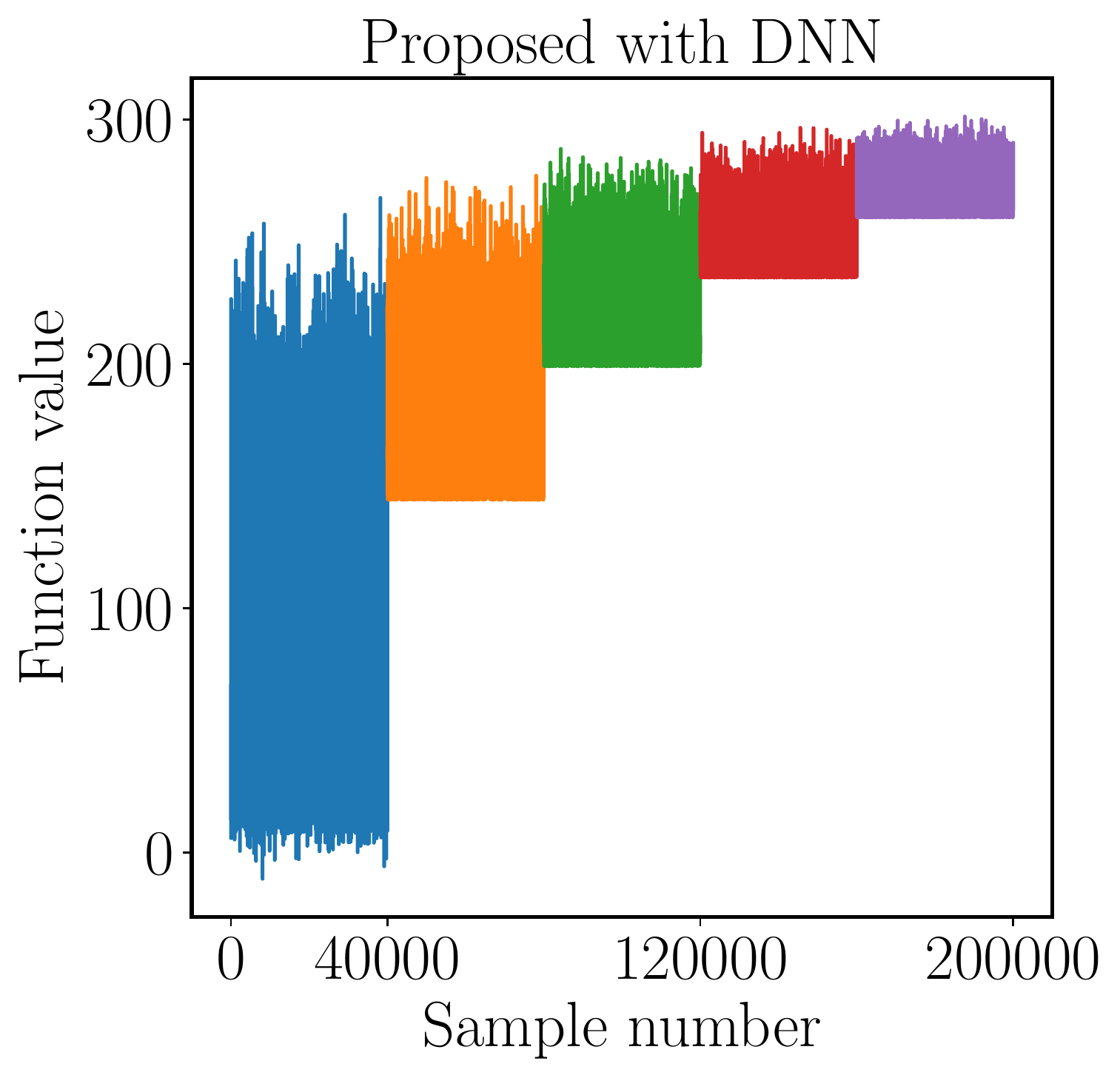} 
\caption{}
\label{Borehole_trace_3}
\end{subfigure}
\caption{Borehole function value trace plot across the five subsets for (a) subset simulation, (b) the proposed algorithm with Gaussian process ($\mathcal{GP}$) as the low-fidelity model, and the (c) proposed algorithm with DNN as the low-fidelity model.}
\label{Borehole_trace}
\end{figure}

Using a subset-independent $U$-function (i.e., Equation \eqref{eqn:u_1}) in the proposed algorithm for this case returns {$P_f=0$ due to the small failure probability}. As discussed in Section \ref{sec:HF_LF}, when the $P_f$ value is small, the function values in the first subset will be far from the required failure threshold $(\mathcal{F})$. Then, since the $\mathcal{GP}$ is trained on a small sample to learn the differences in the HF and LF models, using a subset-independent $U$-function can lead to the problem illustrated in Figure \ref{Failure_Case} {where the samples never approach the true limit surface.} Using subset-dependent $U$-functions (i.e., Equations \eqref{eqn:u_3} and \eqref{eqn:u_3b}) can alleviate this problem. Figure \ref{Borehole_trace_1} presents a function value trace plot for subset simulation. Figures \ref{Borehole_trace_2} and \ref{Borehole_trace_3} present trace plots for the proposed algorithm, using $\mathcal{GP}$ and DNN, respectively, as the LF model {with subset-dependent $U$-functions}. It is observed that using subset-dependent $U$-functions mitigates the problem illustrated in Figure \ref{Failure_Case}, as we can sample from the higher subsets effectively and estimate the $P_f$ value accurately.

\section{Finite element model case studies}

In this section, we apply the proposed framework for active learning with multifidelity modeling to FE model case studies, and evaluate its performance.

\subsection{Steady-state incompressible Navier-Stokes equations}

\begin{figure}[h]
\centering
\includegraphics[width=2.2in, height=2.2in]{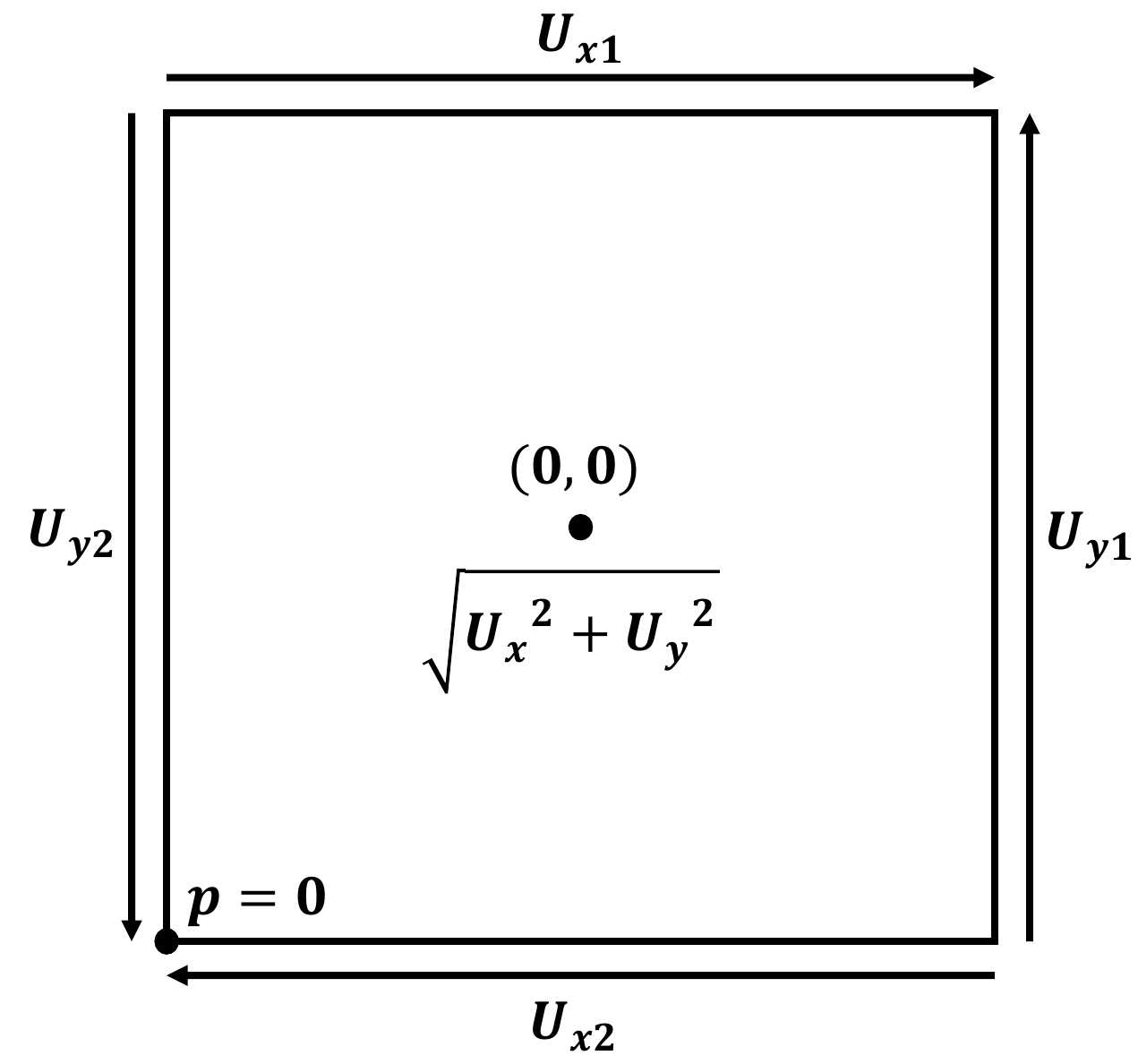} 
\caption{Schematic of the four-sided lid-driven cavity problem. $U_{x1}$, $U_{x2}$, $U_{y1}$, and $U_{y2}$ are the velocities applied along the boundaries of this domain. $\sqrt{U_{x}^2+U_{y}^2}$ is the required resultant velocity at the origin $(0,0)$.}
\label{NS_Domain}
\end{figure}

We consider the four-sided lid-driven cavity problem described in Figure \ref{NS_Domain}. The fluid domain is two dimensional, has random kinematic viscosity ($\nu$) and density ($\rho$), and is subjected to random velocities at the four boundaries. Table \ref{Table:NS_Inputs} describes the variables of this problem along with their probability distributions. We are interested in computing the velocity magnitude at the center of the fluid domain in Figure \ref{NS_Domain}. The HF model solves the Navier-Stokes equations:

\begin{equation}
    \label{eqn:NS_eqn}
    \begin{aligned}
    \frac{1}{\rho} \nabla p + \nabla \cdot (\nu~\nabla \pmb{U}) &= (\pmb{U} \cdot \nabla)~\pmb{U} \\
    \nabla \cdot \pmb{U} &= 0 \\
    \end{aligned}
\end{equation}

\noindent where $p$ is the pressure and $\pmb{U}$ is the velocity vector. The LF model is the Stokes approximation, which ignores the nonlinear convective term in the Navier-Stokes equations:

\begin{equation}
    \label{eqn:Stokes_eqn}
    \begin{aligned}
    \frac{1}{\rho} \nabla p + \nabla \cdot (\nu~\nabla \pmb{U}) &= 0 \\
    \nabla \cdot \pmb{U} &= 0 \\
    \end{aligned}
\end{equation}

\begin{figure}[h]
\begin{subfigure}{0.5\textwidth}
\centering  
\includegraphics[width=2.5in, height=2.5in]{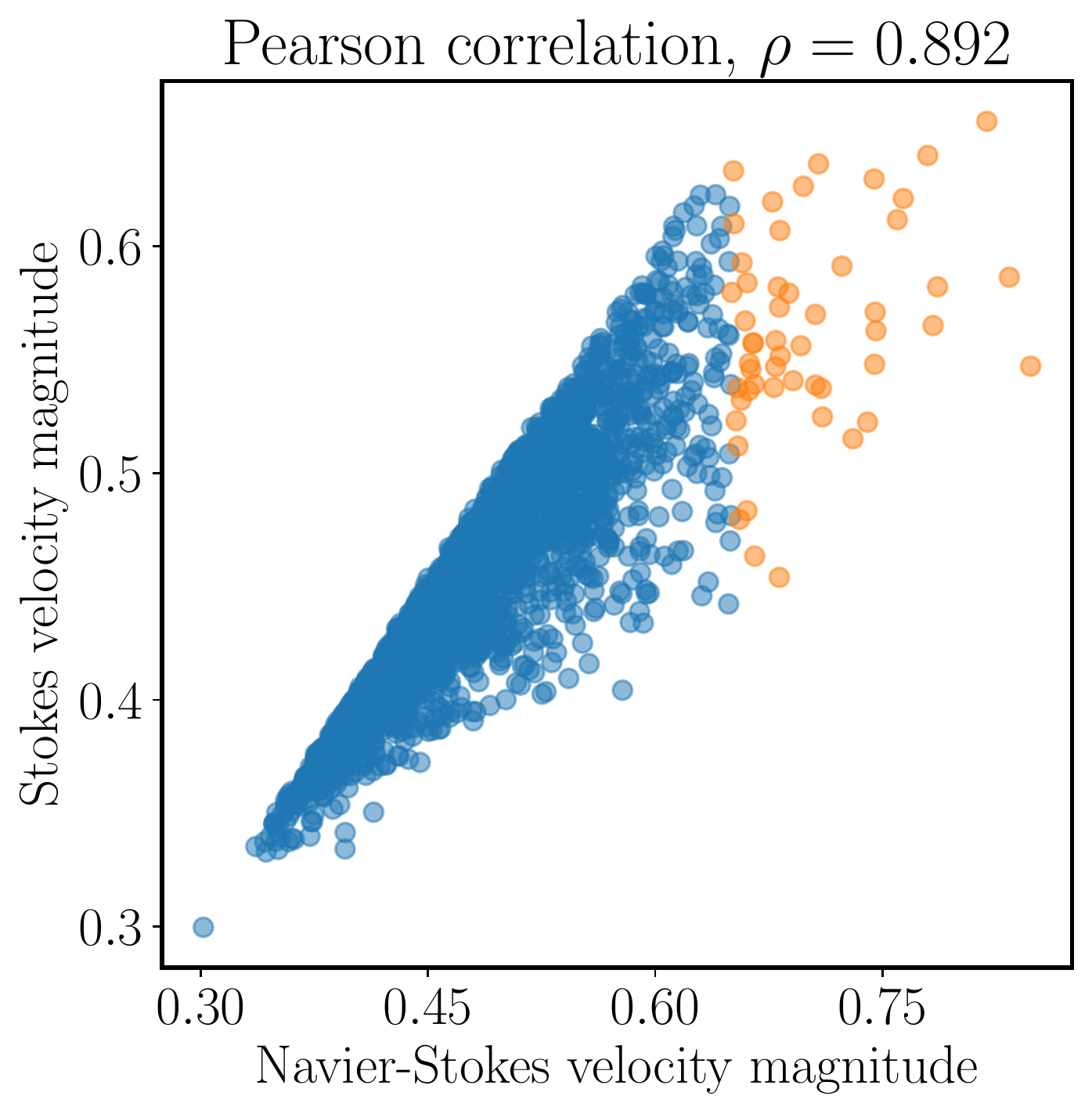} 
\caption{}
\label{NS_S_1}
\end{subfigure}
\begin{subfigure}{0.5\textwidth}
\centering
\includegraphics[width=2.5in, height=2.5in]{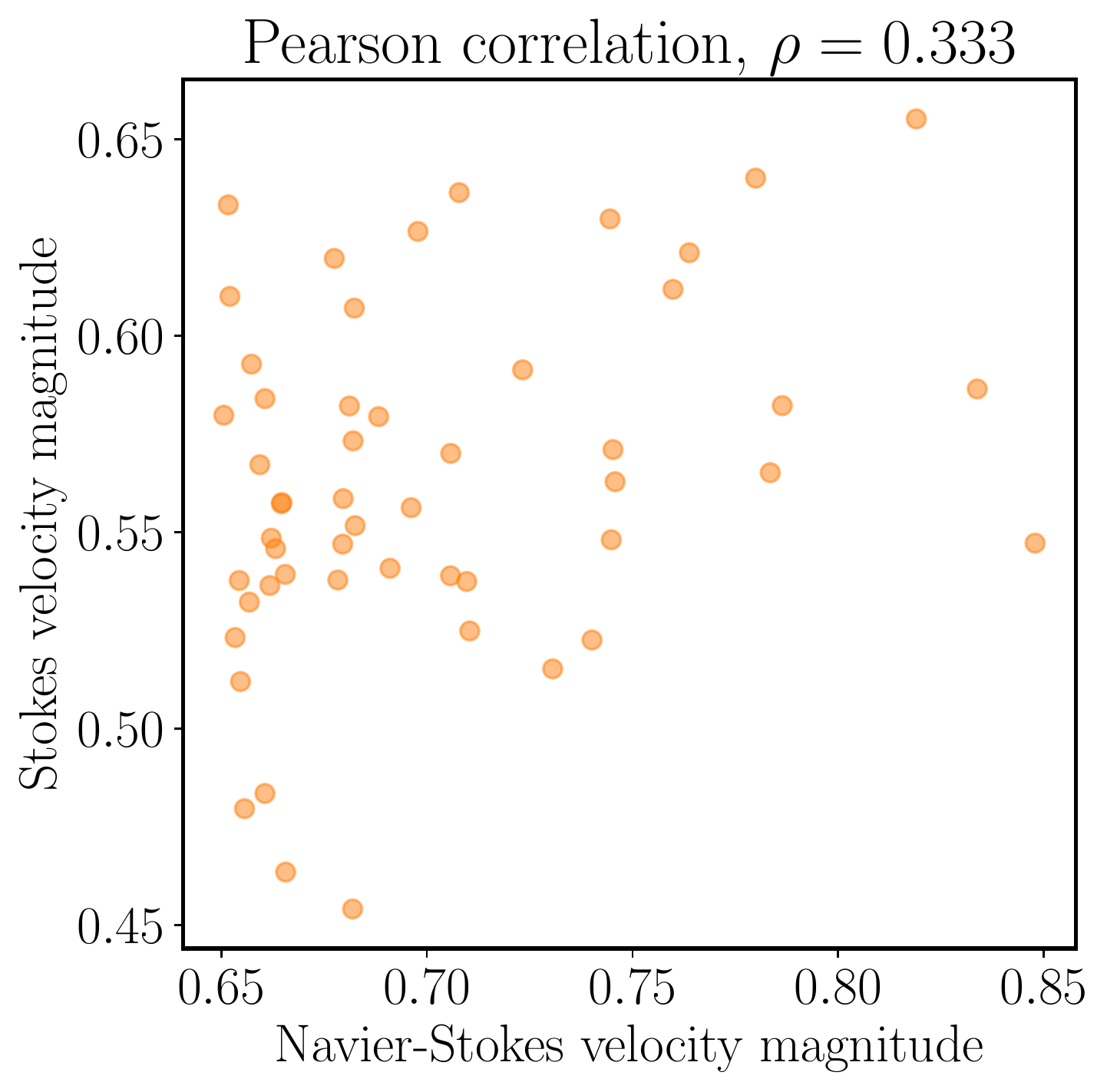} 
\caption{}
\label{NS_S_2}
\end{subfigure}
\caption{(a) Comparison between the Navier-Stokes and Stokes velocity magnitudes for $2,800$ random input samples. The orange dots represent cases in which the Navier-Stokes velocity magnitudes exceed $0.65$. (b) Comparison between the Navier-Stokes and Stokes velocity magnitudes when the Navier-Stokes velocity magnitudes exceed $0.65$.}
\label{NS_S}
\end{figure}

\noindent We solve the HF and LF model equations using the Navier-Stokes module \cite{Peterson2018a} in the Multi-physics Object-Oriented Simulation Environment (MOOSE) \cite{Permann2020a}. Figure \ref{NS_S_1} presents a scatter plot comparing the resultant velocities at the center of the fluid domain, computed using the HF and LF models. This scatter plot was generated using {2,800} evaluations of these models with randomly sampled input variables. The seemingly high overall correlation between HF and LF model velocity magnitudes is due to an abundance of samples with lower velocity magnitudes. For higher velocity magnitudes (i.e., HF model velocity magnitudes of greater than 0.65; represented as orange dots in Figure \ref{NS_S_1}), the correlation decreases {substantially meaning that the LF model is less predictive of the true velocity in this region}. Figure \ref{NS_S_2} presents a scatter plot for the higher velocity magnitudes, and here the correlation is small. 

\begin{table}[h]
\centering
\caption{Parameters in the four-sided lid-driven cavity problem and their probability distributions.}
\label{Table:NS_Inputs}
\small
\begin{tabular}{ |c|c|c|c| }
\hline
\textbf{Variable(s)} & \textbf{Definition} &  \textbf{Distribution} & \textbf{Parameters}\\ 
\hline
$\ln{\nu}$ & Kinematic viscosity & Truncated Normal & \Centerstack[c]{{Mean: $\ln{0.025}$}\\{Std: $0.5$}\\{Lower: $\ln{0.005}$}\\{Upper: $\ln{0.05}$}} \\ 
\hline
$\rho$ & Density & Uniform & $[0.5,~1.5]$ \\ 
\hline
\Centerstack[c]{{$\ln{U_{x1}},~-\ln{U_{x2}}$}\\{$\ln{U_{y1}},~-\ln{U_{y2}}$}} & \Centerstack[c]{{x velocity at top, bottom}\\{y velocity at right, left}}  & Truncated Normal & \Centerstack[c]{{Mean: $\ln{0.75}$}\\{Std: $0.25$}\\{Lower: $\ln{0.5}$}\\{Upper: $\ln{1.5}$}} \\
\hline
\end{tabular}
\end{table}
\normalsize

\begin{figure}[h]
\begin{subfigure}{0.5\textwidth}
\centering  
\includegraphics[width=3.1in, height=2.7in]{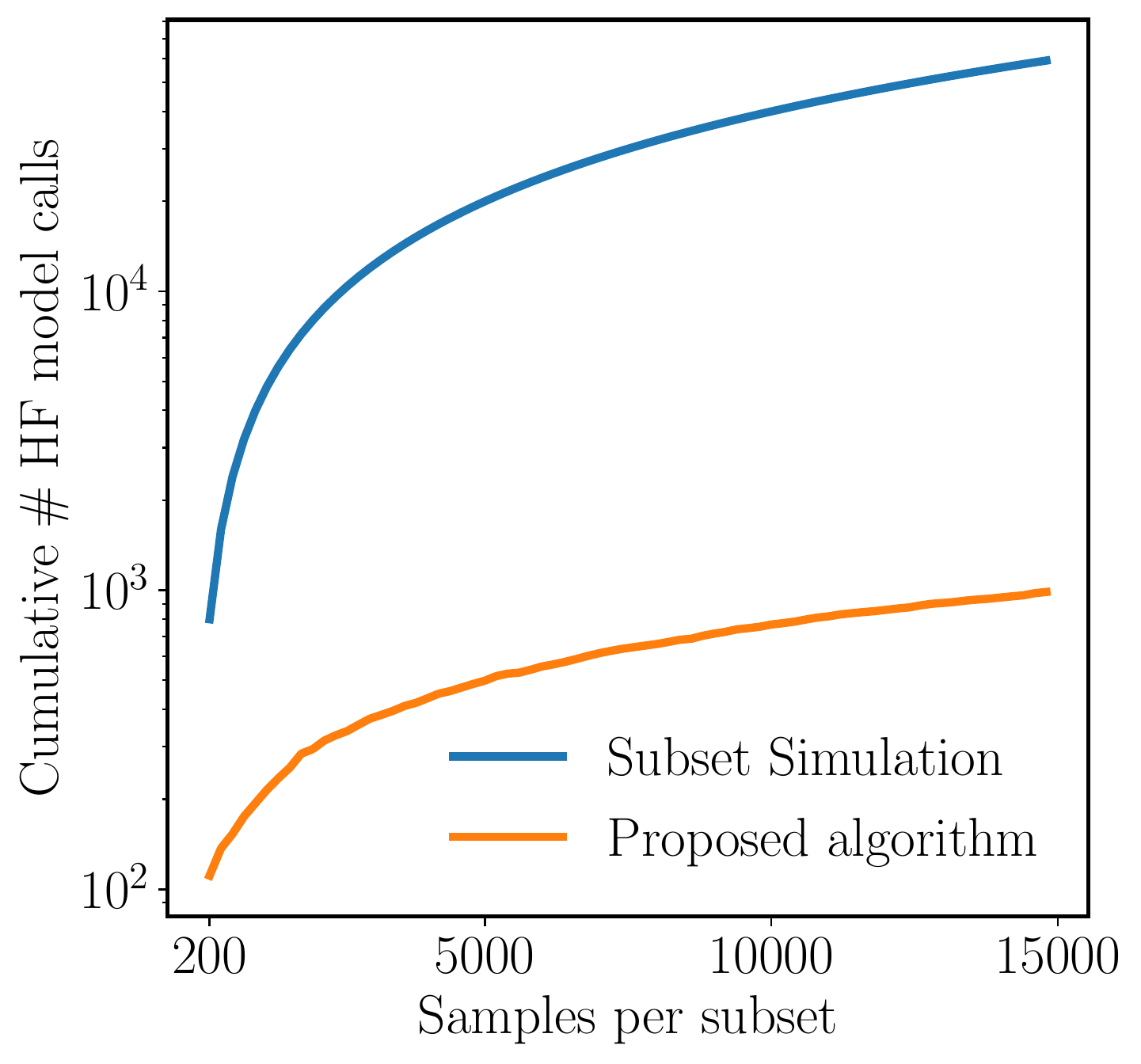} 
\caption{}
\label{NS_convergence_1}
\end{subfigure}
\begin{subfigure}{0.5\textwidth}
\centering
\includegraphics[width=2.9in, height=2.7in]{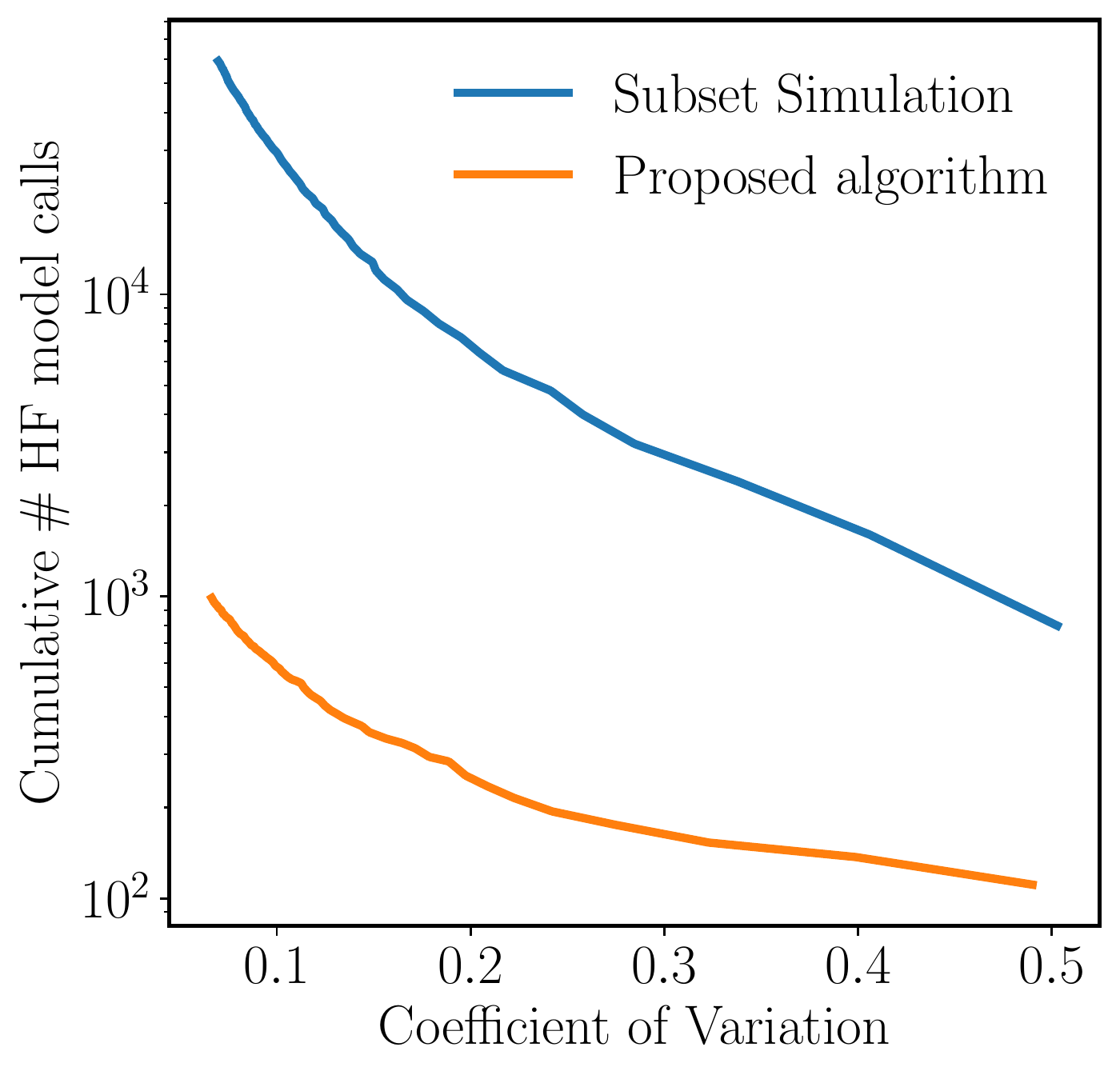} 
\caption{}
\label{NS_convergence_2}
\end{subfigure}
\caption{Cumulative number of calls to the high-fidelity model with (a) the number of samples per subset and (b) the  coefficient of variation for the Navier-Stokes case study.}
\label{NS_convergence}
\end{figure}

The failure threshold chosen for this example is a velocity magnitude $\mathcal{F} = 0.85$. We used 20 evaluations of the HF and LF models to {initially} train the {active learning} $\mathcal{GP}$ in the proposed algorithm to learn their differences. Subset-dependent $U$-functions were used in the proposed algorithm. Table \ref{Table_NS} presents the results computed using the proposed algorithm and subset simulation using the HF model. For similar COVs, the $P_f$ values for both methods are {in close agreement}. Additionally, the proposed algorithm requires only a fraction of calls to the HF model as compared with subset simulation. Figure \ref{NS_convergence_1} presents the cumulative number of HF model calls across all subsets, with respect to the number of samples in each subset, for the three subset-based methods. Figure \ref{NS_convergence_2} presents the cumulative number of HF model calls with {the COV}.

\begin{table}[h]
\centering
\caption{Result comparison between subset simulation and the proposed algorithm in regard to the four-sided lid-driven cavity problem.}
\label{Table_NS}
\small
\begin{tabular}{ |c|c|c| }
\hline
& \Centerstack[c]{\textbf{Subset simulation using}\\\textbf{Navier-Stokes}$^{\dagger}$} & \Centerstack[c]{\textbf{Proposed algorithm with Stokes as LF}\\\textbf{model and Navier-Stokes as HF model}$^{\dagger}$} \\ 
\hline
$\mathbf{P_f}$ & {2.36E-4} & 2.03E-4 \\ 
\hline
\textbf{COV} & 0.069 (0.049$^{\ddagger}$) &  0.066 (0.049$^{\ddagger}$)\\ 
\hline
\textbf{$\#$ HF calls} & 60000 & 997 \\ 
\hline
\end{tabular}
\begin{tablenotes}
\item \normalsize{$^{\dagger}$} \scriptsize{Uses four subsets, with 15,000 samples for each}
\item \normalsize{$^{\ddagger}$} \scriptsize{COV value without considering the cross-correlations in the MCMC samples}
\end{tablenotes}
\end{table}
\normalsize

\subsection{Maximum von Mises stress in a 3-D cylindrical domain}

\begin{figure}[h]
\begin{subfigure}{0.5\textwidth}
\centering  
\includegraphics[width=2.72in, height=2.5in]{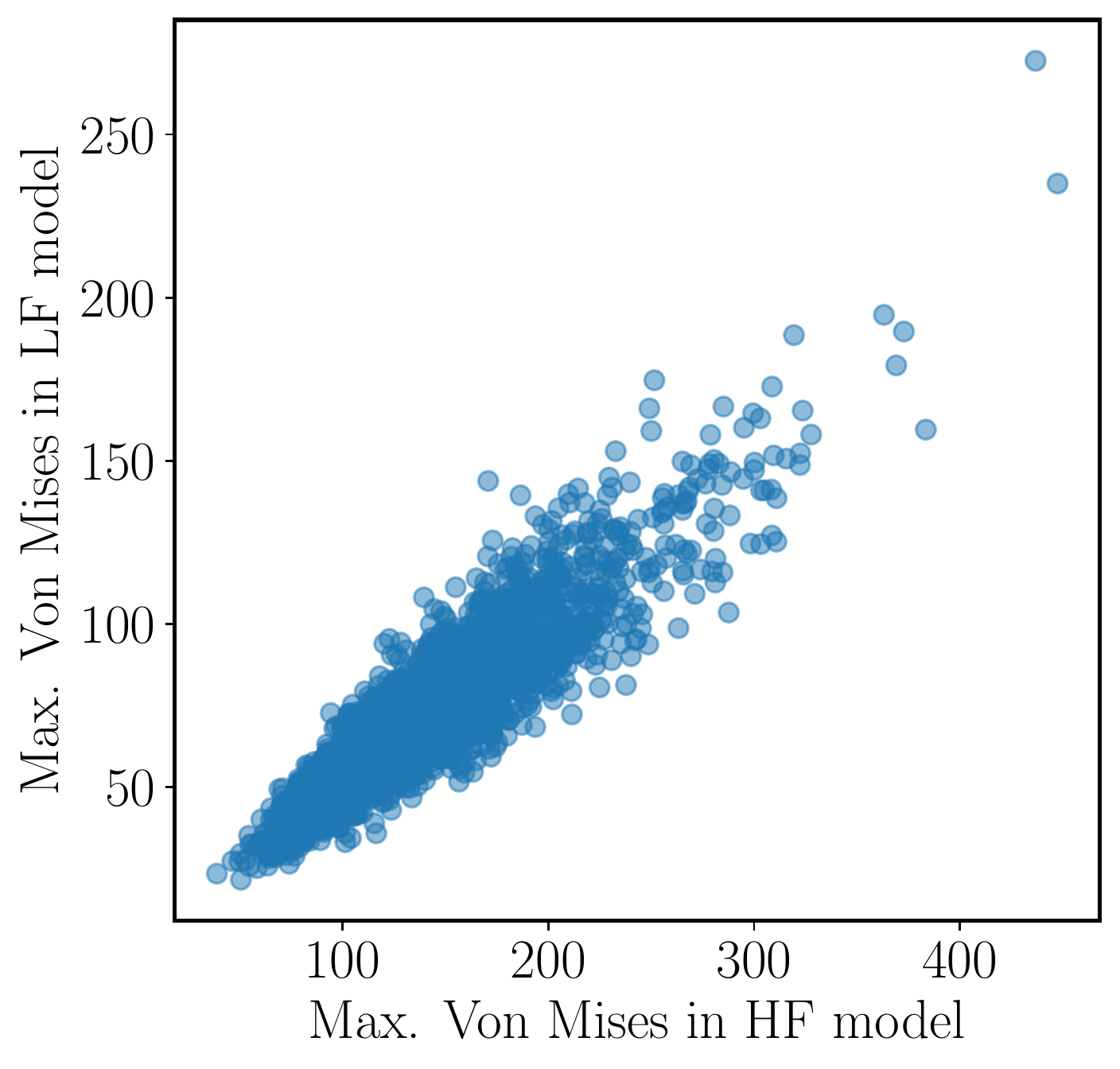} 
\caption{}
\label{Material_S_1}
\end{subfigure}
\begin{subfigure}{0.5\textwidth}
\centering
\includegraphics[width=2.72in, height=2.5in]{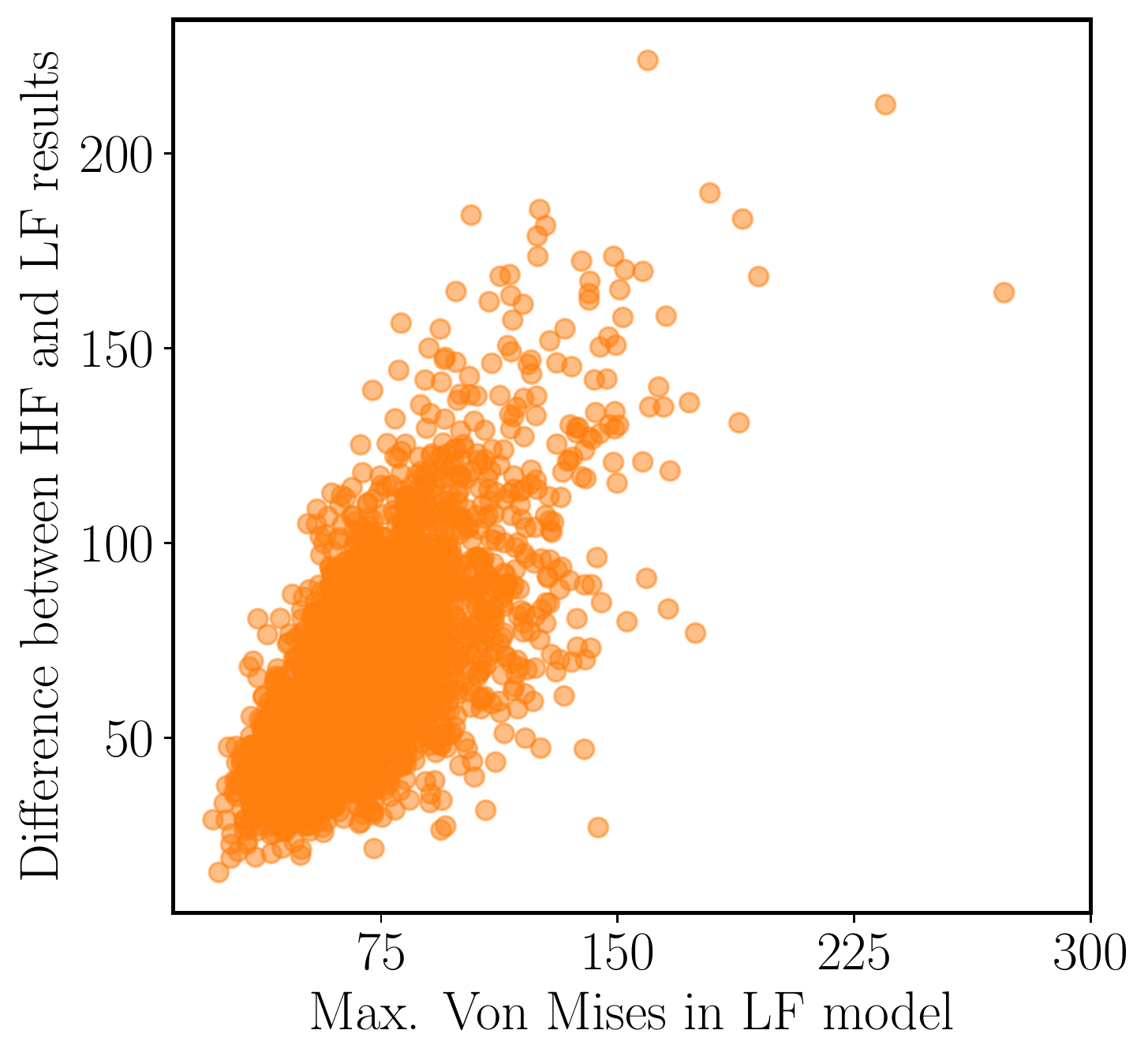} 
\caption{}
\label{Material_S_2}
\end{subfigure}
\caption{(a) Comparison between the von Mises stresses in the high-fidelity and low-fidelity models for $2,800$ random input samples. (b) Comparison between the von Mises stresses in the low-fidelity model and differences in the von Mises stresses in the low- and high-fidelity models for $2,800$ random input samples.}
\label{Material_S}
\end{figure}

We consider a 3-D {solid cylinder} with a radius of 0.5 units and a height of 1 unit. This domain is fixed in all three directions at the bottom end and subjected to random displacements applied to the entire top end in all three directions (i.e., $U_x,~U_y,~U_z$). We are interested in determining the maximum von Mises stress anywhere in the domain. The governing equation{s} for this problem {from continuum} solid mechanics {are}:

\begin{equation}
    \label{eqn:Mat_1}
    \begin{aligned}
    &\widetilde{\nabla} \pmb{\sigma}_s + \mathbf{f}_{bs} = 0\\
    &\textrm{where,}~\widetilde{\nabla} = \begin{bmatrix}
    \frac{\partial}{\partial x_1} & 0 & 0 & \frac{\partial}{\partial x_2} & \frac{\partial}{\partial x_3} & 0\\
    0 & \frac{\partial}{\partial x_2} & 0 & \frac{\partial}{\partial x_1} & 0 & \frac{\partial}{\partial x_3}\\
    0 & 0 & \frac{\partial}{\partial x_3} & 0 & \frac{\partial}{\partial x_1} & \frac{\partial}{\partial x_2}\\
    \end{bmatrix}\\
    \end{aligned}
\end{equation}

\noindent where $\pmb{\sigma}_s$ is the Cauchy stress tensor in Voigt notation and $\mathbf{f}_{bs}$ is the body force vector. The stress-strain relationship is assumed to be linear:

\begin{equation}
    \label{eqn:Mat_2}
    \pmb{\sigma}_s = \mathbf{D}_s~\pmb{\varepsilon}_s
\end{equation}

\noindent where $\pmb{\varepsilon}_s$ is the Cauchy strain tensor in Voigt notation and $\mathbf{D}_s$ is the elasticity tensor. The HF model is transversely isotropic, and its $\mathbf{D}_s$ is defined by five elastic constants $\{E_x,~E_z,~G_{xz},~\nu_{xy},~\nu_{xz}\}$. In the FE solution, the HF model mesh has $10,845$ DoFs. The LF model is isotropic, and its $\mathbf{D}_s$ is defined by two elastic constants $\{E_x,~\nu_{xy}\}$. Additionally, the LF model mesh is coarser, with only $1,608$ DoFs. The elastic constants in both the HF and LF models are treated as random {variables described in} Table \ref{Table:Mat_Inputs}. We solve for the maximum von Mises stress in the HF and LF models using the Tensor Mechanics module in MOOSE \cite{Permann2020a}. Figure \ref{Material_S_1} presents a scatter plot comparing the maximum von Mises stress computed by the HF and LF models. Not only is there a significant scatter between the HF and LF model results, but their scales of the axes are different as well. Additionally, Figure \ref{Material_S_2} presents a scatter plot showing the difference between these HF and LF model results as a function of the LF model result. A significant scatter in this plot is noted, indicating and increased complexity in inferring the ``right'' $\mathcal{GP}$ correction terms in the {proposed algorithm}.



\begin{table}[h]
\centering
\caption{Parameters in the {solid} mechanics problem and their probability distributions.}
\label{Table:Mat_Inputs}
\small
\begin{tabular}{ |c|c|c|c| }
\hline
\textbf{Variable(s)} & \textbf{HF/LF} & \textbf{Distribution} & \textbf{Parameters}\\ 
\hline
$\ln{E_x}$ & HF and LF & Normal & $[\ln{200},~0.1]$ \\ 
\hline
$\ln{E_z}$ & Only HF & Normal & $[\ln{300},~0.1]$ \\ 
\hline
$\ln{\nu_{xy}}$ & HF and LF & Normal & $[\ln{0.25},~0.1]$ \\
\hline
$\ln{\nu_{xz}}$ & Only HF & Normal & $[\ln{0.3},~0.1]$ \\
\hline
$\ln{G_{xz}}$ & Only HF & Normal & $[\ln{135},~0.1]$ \\
\hline
$\ln{U_{x}},~\ln{U_{y}},~\ln{U_{z}}$ & HF and LF & Normal & $[\ln{0.15},~0.5]$ \\
\hline
\end{tabular}
\end{table}
\normalsize

\begin{figure}[h]
\begin{subfigure}{0.5\textwidth}
\centering  
\includegraphics[width=3.1in, height=2.7in]{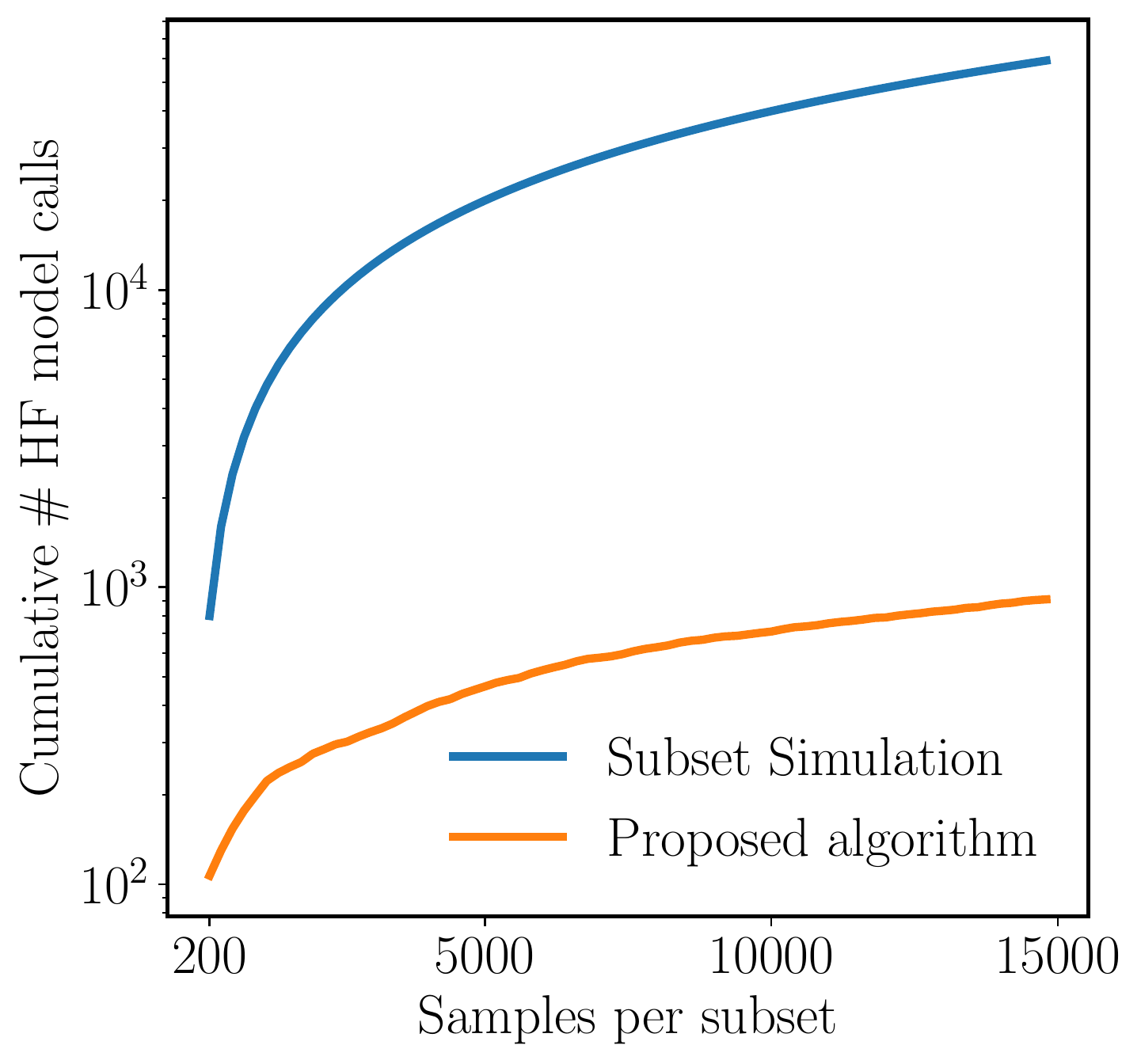} 
\caption{}
\label{Material_convergence_1}
\end{subfigure}
\begin{subfigure}{0.5\textwidth}
\centering
\includegraphics[width=2.9in, height=2.7in]{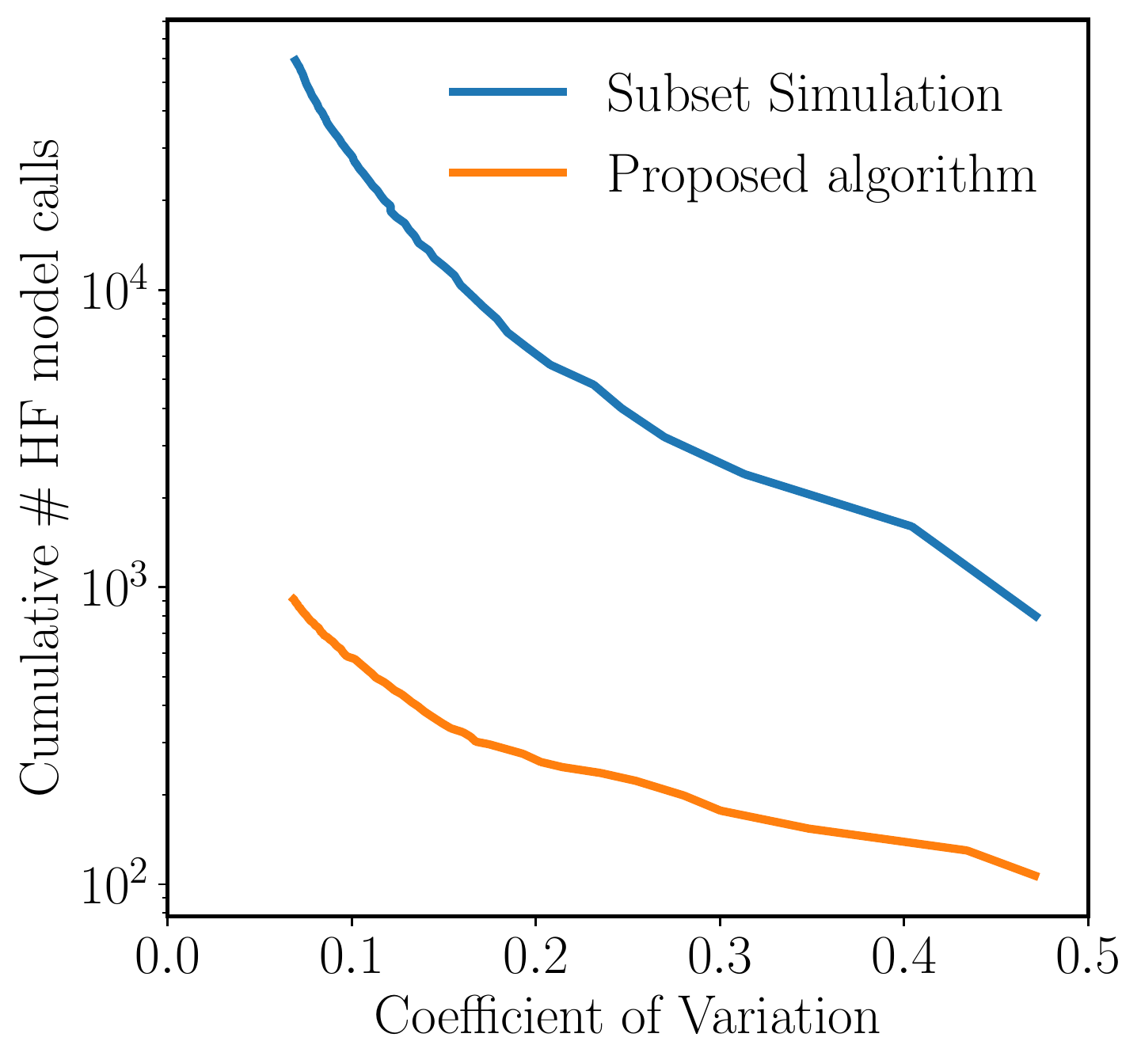} 
\caption{}
\label{Material_convergence_2}
\end{subfigure}
\caption{Cumulative number of calls to the high-fidelity model with (a) the number of samples per subset and (b) the coefficient of variation for the mechanics problem case study.}
\label{Material_convergence}
\end{figure}

The failure threshold assumed here is a the maximum von Mises stress $\mathcal{F} = 400$ MPa. We used 20 evaluations of the HF and LF models to {initially} train the {active learning} $\mathcal{GP}$ in the proposed algorithm to learn their differences. Subset-dependent $U$-functions were used in the proposed algorithm. Table \ref{Table_Material} presents the results computed using the proposed algorithm and subset simulation with the HF model. For similar COV values, the $P_f$ values for both methods are not only close, but the proposed algorithm requires only a fraction of calls to the HF model as compared with subset simulation. Figure \ref{Material_convergence_1} presents the cumulative number of HF model calls across all subsets, with respect to the number of samples in each subset, for the three subset-based methods. Figure \ref{Material_convergence_2} presents the cumulative number of HF model calls with the {COV}.

\begin{table}[h]
\centering
\caption{Results comparison between subset simulation and the proposed algorithm in regard to the mechanics problem.}
\label{Table_Material}
\small
\begin{tabular}{ |c|c|c| }
\hline
& \Centerstack[c]{\textbf{Subset simulation using}\\\textbf{transversely isotropic}$^{\dagger}$} & \Centerstack[c]{\textbf{Proposed algorithm with isotropic coarse}\\\textbf{mesh as LF model and transversely}\\\textbf{isotropic as HF model}$^{\dagger}$} \\ 
\hline
$\mathbf{P_f}$ & 6.42E-4 & 6.6E-4 \\ 
\hline
\textbf{COV} & 0.069 (0.043$^{\ddagger}$) &  0.068 (0.043$^{\ddagger}$)\\ 
\hline
\textbf{$\#$ HF calls} & 60000 & 913 \\ 
\hline
\end{tabular}
\begin{tablenotes}
\item \normalsize{$^{\dagger}$} \scriptsize{Uses four subsets, with 15,000 samples for each}
\item \normalsize{$^{\ddagger}$} \scriptsize{COV value without considering the autocorrelations in the MCMC samples}
\end{tablenotes}
\end{table}
\normalsize

\section{Summary and conclusions}

Rare events estimation has applications across multiple fields (e.g., aerospace systems reliability, critical infrastructure resilience, and nuclear engineering). {But failure probabilities are very computationally expensive to evaluate, requiring a very large number of model evaluations. When only high-fidelity (HF) models can be leveraged for this task, it often becomes intractable. The ability to leverage low-fidelity (LF) models in this setting can overcome this burden.} An adaptive approach {to} multifidelity modeling that {corrects} a LF model, decides when to call the high-fidelity (HF) model, and learns the failure boundary on the fly can provide the flexibility and robustness for rare events estimation using multiple models, while significantly reducing computational costs. Here, we propose {such} a framework. This framework operates by fusing the LF prediction with a Gaussian process correction term, filtering the corrected LF prediction to decide whether to call the HF model and, for enhanced accuracy of subsequent corrections, adapting the Gaussian process correction term after an HF call. In this framework, no assumptions are made as to the quality of the LF model (it can be a poorly trained surrogate model, reduced physics model, or reduced DoF model) or its correlations with the HF model. Dynamic active learning functions are proposed, and these improved the proposed algorithm's robustness for smaller failure probabilities.

We evaluate the performance of our framework using standard academic case studies {in addition to more computationally advanced} FE model case studies such as predicting the Navier-Stokes velocity magnitudes using a Stokes approximation, as well as the von Mises stress in a transversely isotropic material using a coarsely meshed isotropic material. Across these case studies, our proposed framework not only accurately estimates the small failure probability, it also only required a fraction of calls to the HF model, compared to either {Monte Carlo or {conventional} subset simulation}. Future work includes expanding the framework for active learning with multifidelity modeling to consider multiple LF models, and exploring the trade-off between accuracy and computational time across these models.

\section*{Acknowledgment}

This research is supported through the INL Laboratory Directed Research \& Development (LDRD) Program under DOE Idaho Operations Office Contract DE-AC07-05ID14517. This research made use of the resources of the High Performance Computing Center at INL, which is supported by the Office of Nuclear Energy of the U.S. DOE and the Nuclear Science User Facilities under Contract No. DE-AC07-05ID14517.

\bibliography{Bibliography}

\appendix\label{app:A}

\section{Notations}

\begin{table}[H]
\centering
\begin{tabular}{c c c c} 
 $\pmb{X}_{HF}$ & \Centerstack[c]{Input parameters\\for HF model} & $\pmb{X}_{LF}$ & \Centerstack[c]{Input parameters\\for LF model}\\
 $q(\pmb{X})$ & \Centerstack[c]{Input parameters\\distribution} & $F(\pmb{X}_{HF})$ & \Centerstack[c]{High-fidelity\\model prediction}\\
 $f(\pmb{X}_{LF})$ & \Centerstack[c]{Low-fidelity\\model prediction} & $\widetilde{F}(\pmb{X})$ & Required prediction\\
 $\mathcal{F}$ & Failure threshold & $N_s$ & Number of subsets\\
 $N$ & \Centerstack[c]{Number of simulations\\in any subset} & $N_{mc}$ & \Centerstack[c]{Number of Markov chains\\in any subset}\\
 $\mathcal{GP}$ & Gaussian process & $m(.)$ & $\mathcal{GP}$ mean\\
 $k(.,~.)$ & $\mathcal{GP}$ covariance & $U_s$ & \Centerstack[c]{Active learning function\\for subset $s$}\\
 $\mathcal{U}$ & Active learning threshold & $N_{dim}$ & Dimensionality of $\pmb{X}$\\
 $\epsilon$ & \Centerstack[c]{True difference between the\\HF and LF model predictions} & $\hat{\epsilon}$ & \Centerstack[c]{Predicted mean difference between\\the HF and LF model predictions}\\
 $\pmb{S}$ & \Centerstack[c]{Seeds for the current subset\\used in the MCMC scheme} & $\pmb{S}_{\widetilde{F}}$ & \Centerstack[c]{Outputs corresponding to the\\seeds for the current subset}\\
 $p_o$ & \Centerstack[c]{Intermediate conditional probability\\in subset simulation} & $\mathbf{SORT}_{p_o}^{{\tilde{F}}}(.)$ & \Centerstack[c]{Sort function that returns the\\largest $p_o^{\textrm{th}}$ fraction of a vector \\{according to their $\tilde{F}$ values}}\\
 $p(\pmb{X})$ & \Centerstack[c]{Proposal distribution in\\ the MCMC scheme} & $\alpha$ & MCMC acceptance probability\\
 $\pmb{X}^*$ & \Centerstack[c]{Proposed input vector\\ in the MCMC scheme} & $\widetilde{F}^*$ & \Centerstack[c]{Required prediction corresponding\\ to the proposed input vector}\\
$\mathcal{F}_s$ & \Centerstack[c]{Failure threshold for the\\ subset $s$} & $\mathbf{I}(.)$ & \Centerstack[c]{Indicator function}\\
[1ex] 
\end{tabular}
\label{table:Notations}
\end{table}

\end{document}